\def\year{2021}\relax
\documentclass[letterpaper]{article}
\pdfoutput=1
\usepackage{aaai21} 
\usepackage{times} 
\usepackage{helvet} 
\usepackage{courier} 
\usepackage[hyphens]{url} 
\usepackage{graphicx} 
\usepackage{graphicx} 
\usepackage{natbib}
\usepackage{caption}
\usepackage[utf8]{inputenc} 
\usepackage[T1]{fontenc}    

\usepackage[switch]{lineno}  %

\usepackage{xcolor}

\graphicspath{ {./images/} }

\newcommand{\hide}[1]{}

\newcommand{\SPARQL}{\textsc{sparql}}

\newcommand{\cfq}{CFQ}
\newcommand{\ocfq}{\textsc{o-CFQ}}
\newcommand{\ucfq}{\textsc{u-CFQ}}
\newcommand{\xcfq}{\textsc{x-CFQ}}
\newcommand{\lcfq}{\textsc{l-CFQ}}
\newcommand{\ncfq}{\textsc{n-CFQ}}
\newcommand{\bcfq}{\textsc{b-CFQ}}

\newcommand{\starcfq}{\textsc{*-CFQ}}

\newcommand{\featurespan}[1]{\textbf{#1}}

\title{\starcfq{}: Analyzing the Scalability of Machine Learning on a Compositional Task}

\author{%
Dmitry Tsarkov\setcounter{footnote}{1}\thanks{Equal contribution. A description of each author’s contribution is available in Appendix~\ref{app:contributions}.},
Tibor Tihon\setcounter{footnote}{1}\footnotemark{},
Nathan Scales,
Nikola Momchev, \\[0.5ex]
Danila Sinopalnikov,
Nathanael Sch\"arli \\[0.5ex]
}
\affiliations{
Google Research, Brain Team \\[0.5ex]
\small
\texttt{\{tsar,ttihon,nkscales,nikola,sinopalnikov,schaerli\}@google.com}
}

\begin{document}

\maketitle
\begin{abstract}
We present \starcfq{} (``star-\cfq{}''): a suite of large-scale datasets of varying scope based on the \cfq{} semantic parsing benchmark, designed for principled investigation of the scalability of machine learning systems in a realistic compositional task setting. Using this suite, we conduct a series of experiments investigating the ability of Transformers to benefit from increased training size under conditions of fixed computational cost.
We show that compositional generalization remains a challenge at all training sizes, and we show that increasing the scope of natural language leads to consistently higher error rates, which are only partially offset by increased training data.
We further show that while additional training data from a related domain improves the accuracy in data-starved situations, this improvement is limited and diminishes as the distance from the related domain to the target domain increases.

\end{abstract}

\section{Introduction}
\label{sect:introduction}

Intuitively, if you see a lot of examples of natural language questions about TV shows, it ought to also help understand similar syntax in questions about movies, or in questions that refer to both movies and TV shows together. Ideally, the training examples from the related domain should strictly improve performance, not hurt it.
If you can satisfy that property, then you have at least a chance at eventually achieving arbitrarily robust performance across a range of domains, given sufficient training data in aggregate.

How and to what extent current machine learning (ML) approaches can be made to robustly solve natural language understanding (NLU) at the scale of arbitrary natural language across domain -- with or without access to large quantities of training data -- remains, however, an open question.

On one hand, research into the scaling behavior of deep learning systems has found generalization loss to decrease reliably with training size and model size in a power law or related logarithmic relationship across a range of architectures and tasks, from image classification with convolutional neural networks~\citep{cho2015much} to language modeling with Transformers~\citep{rosenfeld2019constructive,kaplan2020scaling,brown2020language}. Recent results in an i.i.d.\ setting show this pattern to persist across many orders of magnitude, with no established upper limit~\citep{kaplan2020scaling}.

At the same time, it has been shown that current ML systems continue to struggle to achieve robust performance in classes of tasks that require compositional generalization 
~\citep{keysers2020} -- an ability that has been argued to be crucial to robust language understanding~\citep{fodor1988connectionism,Lake2018GeneralizationWS,battagliaetal2018relational,hupkes2019compositionality}.

In this paper, we combine these two lines of research by investigating the effect of training size on error rates in the context of a compositional task. Specifically, we derive a suite of extended datasets based on the Compositional Freebase Questions (\cfq{}) semantic parsing benchmark~\citep{keysers2020}.
We then use the compositional structure of each example to construct controlled experiments that measure the error rates when increasing training size in settings requiring compositional generalization and in settings simulating scaling to a broader scope of natural language.
We apply these experiments to analysis of Transformers~\citep{vaswani2017attention} in a setting of fixed computational cost -- that is, of fixed model size and fixed training steps -- and demonstrate key limits to their scalability in this setting.

Our contributions are the following:

\begin{itemize}
    \item We present \starcfq{} (``star-\cfq{}''): a suite of large-scale datasets and corresponding canonical splits, designed to enable principled investigation of the scalability of ML systems in a realistic compositional task setting.
    The datasets and splits follow the same setup as the original \cfq{}, but span a range of data sizes, rule scopes, and compound divergences with the largest consisting of 76\% more rules, 41x as many examples, 14x as many question patterns, 53x as many \SPARQL{} patterns, and covering a 32x larger domain than \cfq{} (Section~\ref{sect:datasets}).
    \item We confirm that under conditions of fixed computational cost, error rates for Transformers plateau at large training sizes. Moreover, targeting larger scopes of natural language leads to consistently higher error rates which are only partially offset by increased training data (Section~\ref{sect:effect-rule-scope}).
    \item We demonstrate that compositional generalization remains a challenge at all training sizes,
    but that increases in training size continue to reap benefits in situations requiring compositional generalization, even when error rates would seem to have plateaued in the i.i.d. case
    (Section~\ref{sect:effect-compound-divergence}).
    \item We show that while access to additional training data from a related domain improves accuracy in data-starved situations, this improvement is limited and diminishes as the distance from the related domain to the target domain increases
    (Section~\ref{sect:effect-related-domain}).
\end{itemize}

\section{CFQ Benchmark}
\label{sect:cfq}
\citet{keysers2020} introduced the Compositional Freebase Questions (CFQ), which is a simple but realistic and large natural language dataset that is specifically designed to measure compositional generalization.
The task of interest is semantic parsing from a natural language question (such as 'Which art director of [Stepping Sisters 1932] was a parent of [Imre Sándorházi]?') to a SPARQL query, which can then be executed against the Freebase knowledge base. Named entities are anonymized, which is standard practice and ensures that the models do not have to learn all the entities.

CFQ was constructed using the Distribution-Based Compositionality Assessment (DBCA) method. This means that the dataset is automatically generated from a set of rules in a way that precisely tracks which rules (atoms) and rule combinations (compounds) were used to generate each example. 
Using this information, the authors generate ``maximum compound divergence'' (MCD) splits, which maximize the compound divergence while guaranteeing a small atom divergence between train and test sets. MCD splits are well suited for assessing compositionality because they are both fair (because the distribution of individual rules is similar) and compositionally challenging (because the distribution of compounds is as different as possible).

The authors release a number of MCD splits for CFQ, and show that there is a strong negative correlation between the accuracy of three standard sequence-to-sequence architectures and the compound divergence. They investigate this for \textit{LSTM+attention} (an LSTM \citep{hochreiter1997long} with attention mechanism \citep{bahdanau-attention-iclr}), for \textit{Transformer} \citep{vaswani2017attention}, and for \textit{Universal Transformer} \citep{dehghani2018universal}.

In a follow-up publication, \citet{furrer2020compositional} show that this negative correlation also applies to architectures that specifically target compositional generalization (such as CGPS~\citep{li2019compositional} and Neural Shuffle-Exchange Networks~\citep{nsen}) and cannot be overcome by masked language model pre-training using the Text-to-Text Transfer Transformer (T5)~\citep{raffel2019exploring}.

\section{\starcfq{}}
\label{sect:datasets}

We present here \textbf{\starcfq{}}\footnote{Available soon at \url{ https://github.com/google-research/google-research/tree/master/star_cfq}}, a suite of datasets building on the same overall structure and base rule set as \cfq{}, but with two key differences intended to facilitate investigation of the scalability of solutions to the semantic parsing task:

\begin{itemize}
\item \textbf{Increased data size:} The datasets span a range of data sizes, with the largest
41x
the size of \cfq{}.
\item \textbf{Expanded rule set:} The datasets span a range of rules scopes, based on grammar extensions to support additional Freebase types and properties (via new \textit{leaf rules}) and additional syntactic constructs (via \textit{non-leaf rules}).
\end{itemize}

All datasets in the suite include detailed instrumentation of the compositional structure of each example, in a similar format to that used in CFQ.



\subsection{Increased Data Size}
\label{sect:increased-data-size}

The \starcfq{} datasets are generated following the algorithm described in \citet{keysers2020} using rule sets closely related to the \cfq{} rules, but with sampling run at a larger scale in order to generate significantly larger datasets. As in \citet{keysers2020}, after the initial sampling phase, we apply sub-sampling and then semantic and structural filtering to increase the diversity of rule combinations while maintaining a balance of complexity levels and reducing the number of unnatural-sounding questions. The one difference from \citet{keysers2020} is that in order to avoid an observed performance bottleneck, we omit the step of grounding the question in Freebase, which means that the generated questions contain only entity placeholders, without the guarantee that an actual set of entities can be found in Freebase that would lead to a non-empty answer to the question. In Appendix~\ref{app:ungrounded}, we show that while omitting the grounding step leads to a higher incidence of semantically implausible questions, the behavior of the baseline ML systems are highly consistent between the grounded and ungrounded datasets, which motivates our choice to use ungrounded datasets as proxies for grounded ones when exploring the behavior of ML systems at larger scales of training data.


We apply this procedure first to a nearly identical rule set as \cfq{} to generate the large ungrounded dataset \textbf{\ucfq{}} (see Appendix~\ref{app:ucfq-rule-set} for details).
Datasets generated by the same procedure applied to different rule sets are described below in Section~\ref{sect:rule-set-lattice}.
Table~\ref{tab:xcfq-size} shows summary size statistics of two datasets from \starcfq{} compared with \cfq{}. The size statistics for all the \starcfq{} datasets can be found in Appendix~\ref{app:datasets}.


\begin{table}[tb]
    \begin{tabular}{@{}lrrr@{}}
        \hline
        Dataset Statistics & \cfq{} & \ucfq{} & \xcfq{} \\
        \hline
        \hline
        Unique questions & 239,357 & 9,925,221 & 9,879,894 \\  
        Question patterns & 49,320 & 319,407 & 713,137 \\  
        Unique queries & 228,149 & 6,551,678 & 7,249,705 \\   
        Query patterns & 34,921 & 762,680 & 1,847,555 \\ 
        \hline
        Open questions & 108,786 & 4,658,177 & 3,879,020 \\
        Closed questions & 130,571 & 5,267,044 & 6,000,814 \\
        \hline
    \end{tabular}
    \centering
    \caption{Statistics of the largest of the \starcfq{} datasets, in comparison with the original \cfq{}. ``Question pattern'' here corresponds to ``Question patterns (mod entities, verbs, etc.)'' from \citet{keysers2020}, while ``Query pattern'' corresponds to ``Query patterns (mod entities and properties)''.}
    \label{tab:xcfq-size}
\end{table}

\subsection{Extended Rule Set}
\label{sect:extended-rule-set}

In order to simulate coverage of a greater scope of natural language, we enrich the \cfq{} grammar to include up to
$92\%$
more \textit{leaf rules}, which provide support for additional Freebase types and properties or add new surface forms, and up to $37\%$ more \textit{non-leaf rules}, which provide support for additional syntactic constructs. 
Examples of newly supported questions are presented in Table~\ref{tab:grammar-features}. More details on the new language features can be found in Appendix~\ref{app:lattice}.

\begin{table}[tb]
    \begin{tabular}{l@{}}
        \hline
        \featurespan{In} which TV program did M1 \featurespan{play} \\
        Did M0 \featurespan{direct M2 and marry M1} \\
        Who was a \featurespan{crime fiction film}'s Indian writer \\
        What was M1's Anglican art director's \featurespan{ethnicity} \\
        Was the \featurespan{daughter} of the \featurespan{brother} of M0 M2 \\
        \hline
    \end{tabular}
    \centering
    \caption{Examples of newly supported questions.}
    \label{tab:grammar-features}
\end{table}

\subsection{Rule Set Lattice}
\label{sect:rule-set-lattice}

As we are interested in exploring the effect of both the number of new rules added and their type (leaf vs.\ non-leaf), we prepare a suite of rule sets with varying number of leaf and non-leaf rules, which together form a \textit{rule set lattice}.
We then generate a separate large-scale dataset corresponding to each rule set in the lattice. For convenience, we will use the same name (e.g., \bcfq{}, \xcfq{}, etc.) to refer to both the rule set and the largest dataset generated from that rule set.

At the bottom of the lattice is a base rule set which we name \textbf{\bcfq{}}, containing the rules shared by all other rule sets in the lattice. This rule set is designed to be as close as possible to \ucfq{}, but with some minimal adjustments to enable separate evaluation of the addition of leaf- vs.\  non-leaf rules (see Appendix~\ref{app:rule-sets} for details).

\textbf{\lcfq{}} is a rule set containing the rules of \bcfq{} plus all additional leaf rules. \textbf{\ncfq{}} consists of the rules of \bcfq{} plus all additional non-leaf rules. \textbf{\xcfq{}} contains the union of rules from \lcfq{} and \ncfq{}.

In order to test the effect of adding varying numbers of rules of a similar type, we also provide rule sets \textbf{half-\lcfq{}}, \textbf{half-\ncfq{}}, and \textbf{half-\xcfq{}}, which have only half of the additional rules of the relevant types added.

The main characteristics of the rule sets from the lattice are shown in Table~\ref{tab:rule-sets}, with more details in Appendix~\ref{app:lattice}.

\begin{table*}[tb]
    \begin{tabular}{@{}llcccc@{}}
    \hline
         Rule set & Description & \multicolumn{3}{c}{\# Grammar rules} & \# All rules\\
         & & (Non-leaf) & (Leaf) & (Total) & \\
        \hline \hline 
        \cfq{} & Original \cfq{} rule set from \citet{keysers2020} & 54 & 157 & 211 & 443 \\
        \hline
        \ucfq{} & Rule set for ungrounded version of \cfq{} & 54 & 158 & 212 & 444 \\
        \hline
        \bcfq{} & Base \cfq{} & 55 & 137 & 192 & 412 \\
        \hline
        \ncfq{} & Base \cfq{} + additional non-leaf rules & 70 & 139 & 209 & 455 \\
        half-\ncfq{} & Base \cfq{} + half the additional non-leaf rules & 62 & 139 & 201 & 447\\
        \hline
        \lcfq{} & Base \cfq{} + additional leaf rules & 59 & 324 & 383 & 738 \\
        half-\lcfq{} & Base \cfq{} + half the additional leaf rules & 59 & 236 & 295 & 597\\
        \hline
        \xcfq{} & Base \cfq{} + both types of additional rules & 74 & 331 & 405 & 799 \\
        half-\xcfq{} & Base \cfq{} + half of both types of additional rules & 66 & 238 & 304 & 602\\
        \hline
    \end{tabular}
    \centering
    \caption{A list of the new \cfq{}-based rule sets with characteristics.}
    \label{tab:rule-sets}
\end{table*}

\section{Experiments on Effect of Rule Scope}
\label{sect:effect-rule-scope}

\emph{Increased rule scope yields consistently higher error rates, which are only partially offset by increased
training size.}\\

\begin{figure}[tb]
   \centering
   \includegraphics[width=0.47\textwidth]{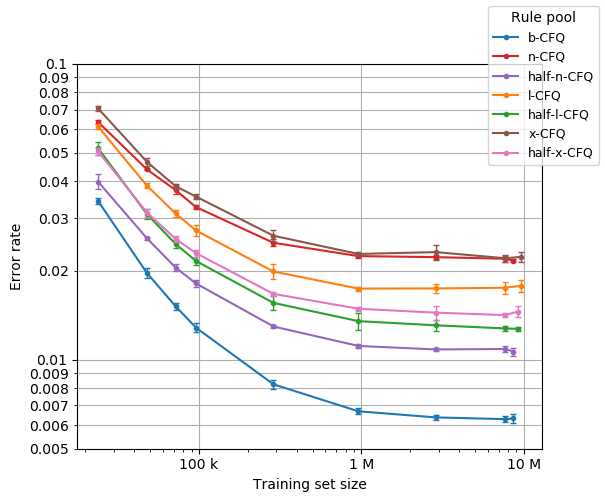}
   \caption{Effect of rule scope on scalability curve. Error rate vs train set size, all datasets from the lattice, double log scale. Every point here (and in all other graphs with random splits) averages values from 5 replicas, error bars represent error margins for confidence level 0.95.
   }
   \label{fig:effect-rule-scope}
\end{figure}

\noindent
Figure~\ref{fig:effect-rule-scope} plots error rates vs.\ training size for a Transformer evaluated on random splits of datasets of increasing rule set scope. In each experiment, the model size and number of training steps (and hence, computational cost) are held constant, with hyperparameters as described in Appendix~\ref{app:hyperparameters}.

Previous research has shown a power law relationship between data size and error rates or loss across a variety of deep learning architectures and tasks when model size and data size are increased in tandem~\citep{kaplan2020scaling, brown2020language, rosenfeld2019constructive, hestness2017deep}.

The results shown in Figure~\ref{fig:effect-rule-scope} are compatible with this previous research in that for all rule scopes, error rates initially vary in a rough power law relationship with training size, while flattening out at higher training sizes. We suspect that the flattening-out is due primarily to the fixed computational cost setting, and that with sufficiently large model sizes the power law relationship would persist longer.

What is most notable in Figure~\ref{fig:effect-rule-scope} is that as the rule scope increases, the asymptotic value of the error rate increases consistently, such that \xcfq{} plateaus at an error rate roughly 3 times that of \bcfq{}. This suggests a notable scalability implication for the Transformer architecture, in that, despite the fact that every dataset in the \starcfq{} suite is fully described by a small set of rules, the moderately-sized Transformer fails to fully capture the rules regardless of the number of training examples provided, and training data alone is insufficient to fully offset the increased error rates that result from even modest increases to the rule scope.

It is also notable that adding non-leaf rules impacts error rate more than adding leaf rules, as evidenced by the error rates for \ncfq{} being considerably higher than \lcfq{} -- in fact, nearly matching the error rates of the combined rule set \xcfq{}. As shown in Appendix~\ref{app:domain-size}, this is despite the fact that the \lcfq{} rule set describes a much larger space of possible questions than that described by \ncfq{}.
This suggests that Transformers struggle more with generalizing to new complex rule combinations than with generalizing to ``primitive substitutions''~\citep{li2019compositional,russin2019comp,Gordon2020Permutation}, in which one leaf element is replaced by another within a question pattern observed in training.

\section{Experiments on Effect of Compound Divergence}
\label{sect:effect-compound-divergence}

\emph{Compound divergence dramatically affects error rates at all training sizes.
Large increases in training data yield greater benefit, however, at high compound divergence than at low.
}\\

\noindent
\citet{keysers2020} observe that as the compound divergence between a train set and test set increases -- that is, as the need for compositional generalization increases -- the accuracy of a Transformer on the \cfq{} task decreases dramatically from over 98\% at compound divergence 0 (a random split) to less than 20\% at compound divergence 0.7 when training size is around 100k examples. In their Appendix H they also present preliminary results of the effect of training size on this performance gap -- specifically, that the performance gap is even wider at smaller training sizes (e.g., around 10k examples) and shrinks as training size increases. This leaves open the question as to whether further increasing the training size beyond 100k could at some point narrow the performance gap sufficiently that the systems can be said to have effectively learned to compositionally generalize.

\begin{figure}[tb]
   \centering
   \includegraphics[width=0.47\textwidth]{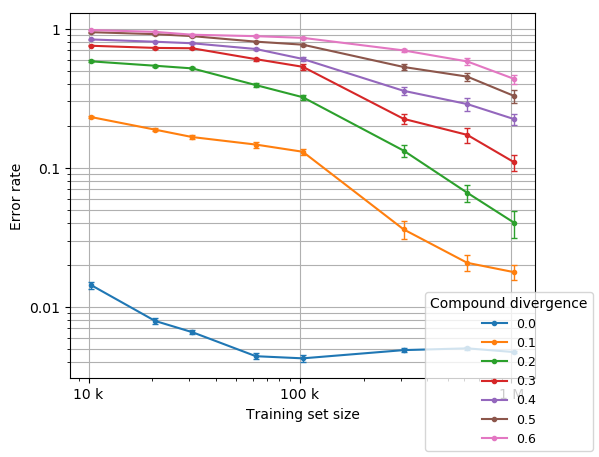}
   \caption{Effect of compound divergence on scalability curve.
   Error rate vs train set size for compound divergence splits for \ucfq{}. Every point averages 5 replicas of up to 36 splits, generated using different random seeds; error bars represent error margins for confidence level 0.95.
   }
   \label{fig:experiment-3.2}
\end{figure}

We investigate this question by reproducing the experiments of \citet{keysers2020} across a wider range of training sizes, from around 10k up to nearly 900k examples. As seen in the results in Figure~\ref{fig:experiment-3.2}, compositional generalization remains a challenge,
with even a moderate compound divergence of 0.2 yielding error rates an order of magnitude higher than those at compound divergence 0 even at the largest training sizes.
Similarly to results on random splits, error rates vary in a rough power law relationship with training size in ranges of moderately large training data, while presumably plateauing at very large training sizes. However, a noticeable warm-up period can be observed, in which error rates decrease at a much slower rate, with the warm-up period persisting longer, the greater the compound divergence. Also, while in Figure~\ref{fig:effect-rule-scope} error rates on random splits consistently began to flatten out starting at around 100-300k training examples (with the plateau even more pronounced in Figure~\ref{fig:experiment-3.2} for the compound divergence 0 split), at the higher compound divergence levels, the plateau is yet to be seen for training sizes up to 1M. This suggests that greater benefit can be reaped from very large training sizes in scenarios requiring compositional generalization than in i.i.d. settings. Further experiments at larger training sizes would be required to verify how much compound divergence affects the final asymptotic error level.

A more detailed comparison of our investigation with that of \citet{keysers2020} is covered in Appendix~\ref{app:divergence-comparison-with-prev-paper}.

\section{Experiments on Effect of Data from Related Domain}
\label{sect:effect-related-domain}

While the experiment results from Section~\ref{sect:effect-rule-scope} show a clear trend of increasing error rate as rule scope increases, this trend can be seen as the combined effect of multiple simultaneously changing factors, which we may group roughly into \textit{test set effects} and \textit{train set effects}.

While increasing rule scope affects the test set on one hand
in the obvious way of increasing its breadth,
there may be more subtle effects that could lead to either an increase or decrease in inherently ``easy'' or ``hard'' examples in the test set, depending on which existing rules the newly added rules tend to compose most easily with.
From the perspective of investigating the scalability of ML systems, these test set effects are of limited interest, as they largely represent artifacts of the specific data generation and sampling algorithm.

Of greater interest are the train set effects. On one hand, we expect increasing the rule scope to \emph{dilute} the train set, in that, from the perspective of any given test example, the proportion of train examples directly relevant to it (e.g., which use a large fraction of the rules or combinations of rules that appear in it) will decrease as the rule scope covered by the train set increases.
At the same time, however, increasing the rule scope may increase the diversity of contexts in which each rule is seen, which may improve generalization~\citep{hill2019environmental}. A key question governing how well ML systems can scale to larger scopes of natural language is indeed how much benefit the system can derive from these training examples that are only partially or indirectly relevant to any given
subset of test examples.

To better answer this question while minimizing interference from test set effects, we shift our focus to the following experiment setup. We choose as point of reference a fixed test set $W_{TEST}$, which is randomly sampled from some limited-scope \emph{target domain} $W$ -- specifically, here we will use the \bcfq{} dataset.
We then suppose we have access to a limited set $W_{TRAIN}$ of $n_{in}$ training examples drawn from the identical distribution as $W_{TEST}$, plus a potentially larger supplementary set $V_{TRAIN}$ of $n_{sup}$ training examples sampled from a \emph{related domain} $V$ that is related to $W$ through some overlap in the rule sets used to generate them. $V$ might consist, for example, of other movie-related questions that use some different syntactic constructs (i.e., different non-leaf rules) that are not present in $W$, or of questions that follow a similar syntax but include words referring to different Freebase properties and types (i.e. different leaf rules). We investigate, for various choices of supplementary domain $V$ and values of $n_{in}$ and $n_{sup}$, the degree to which blending the supplementary data set $V_{TRAIN}$ into the train set improves or harms performance on the original test set $W_{TEST}$.


In each case, we use as supplementary domain $V$ one of the datasets from the \starcfq{} rule set lattice -- specifically one of half-\lcfq{}, \lcfq{}, half-\xcfq{} or \xcfq{} -- with examples generatable by the \bcfq{} rule set filtered out. We omit experiments involving half-\ncfq{} or \ncfq{}, which do not lend themselves well to this experiment setup, due to the high overlap between their domains and \bcfq{}. (See Appendix~\ref{app:blending-with-ncfq} for details.)

For all the experiments we use an approach similar to the one in \citet{keysers2020}:
\begin{itemize}
    \item We use a Transformer architecture with hyperparameters described in Appendix~\ref{app:hyperparameters}.
    \item For each split we train 5 replicas and average the results.
\end{itemize}

Details of how the train sets $W_{TRAIN}$ and $V_{TRAIN}$ are blended are described in Appendix~\ref{app:blending-algorithm}.

In all experiments we used a fixed test set of $95,\!742$ examples sampled randomly from \bcfq{}.

\subsection{Equal Weighting of Examples from Target and Related Domains}
\label{sect:equal-weighting}

\emph{Transformers are sensitive to the training distribution, not just the information contained in the training data. When highly data-starved, adding training examples from a related domain can improve performance. As training size increases, however, the skew in train vs.\ test distribution resulting from expansion of the train set quickly outweighs any benefits from the additional training examples.}\\

In this experiment, we weight examples from $W_{TRAIN}$ and $V_{TRAIN}$ equally, so that as the number of examples in $V_{TRAIN}$ increases, we can observe the same dynamics we would see when scaling to increasing scopes of language, where expanding the scope of the train set simultaneously increases the amount of total information in the train set while also diluting the fraction of the train set that is directly relevant to any given slice of test examples.

\begin{figure*}[tb]
  \centering \footnotesize
  (a)\includegraphics[width=0.47\textwidth]{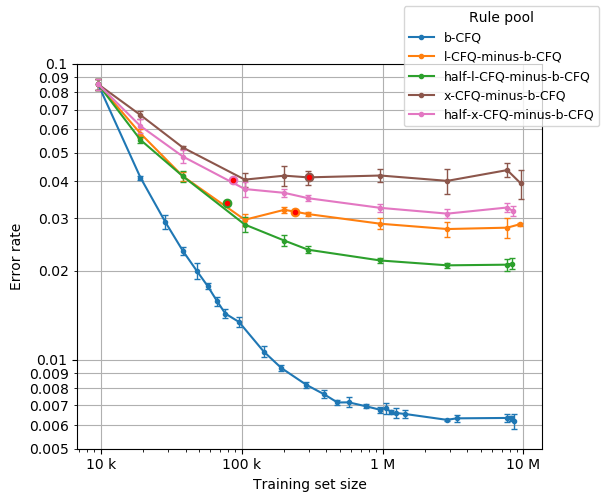}
  (b)\includegraphics[width=0.47\textwidth]{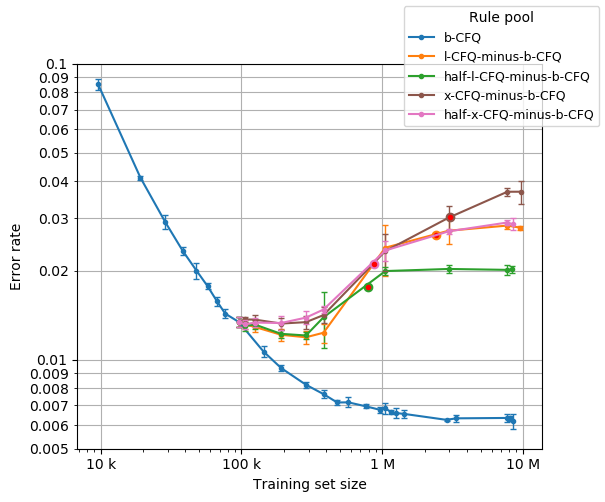}
  \caption{Error rate vs train set size, log-log scale, equal weighting of examples from target and related domains: \textbf{10k} examples (a) and \textbf{100k} examples (b) , \bcfq{} test set, \bcfq{} + others train set.}
  \label{fig:experiment-6-8-fixed}
  \label{fig:experiment-6-8-fixed-small} 
  \label{fig:experiment-6-8-fixed-large} 
\end{figure*}

Figure~\ref{fig:experiment-6-8-fixed-small}(a) summarizes the results when $n_{in}$ is fixed at 10k examples and $n_{sup}$ varies between 10k and 8M.

As in Figure~\ref{fig:effect-rule-scope}, error rate varies with training size in a clear power law relation for training sizes up to around 100k-300k (depending on the dataset), after which performance plateaus. However, the more distant the related domain is from the target domain in terms of number of leaf or non-leaf rules added, the more slowly the error rates drop, and the higher the asymptotic value at which they plateau. This reinforces that the increased error rates observed in Section~\ref{sect:effect-rule-scope} for increased rule scope are indeed driven largely by the changing composition of the train set, rather than purely by test set effects.

For easier comparison with Figure~\ref{fig:effect-rule-scope}, we mark on each curve the point at which the relative size of $n_{sup}$ vs.\ $n_{in}$ matches the ratio that would be expected based on the relative sizes of the domains of $V$ vs.\ $W$, if the train set were sampled from $V$ as a whole, while the test set were simply fixed to observe the performance on the subset of $V$ generated by the narrower rule set $W$.
It can be seen that the error rates at the marked points in Figure~\ref{fig:experiment-6-8-fixed-small} are slightly higher than at similar training sizes in Figure~\ref{fig:effect-rule-scope}, suggesting that a small degree of test set effect is also 
at play.

Figure~\ref{fig:experiment-6-8-fixed-large}(b) summarizes the results when $n_{in}$ is fixed at the larger value of 100k examples, while $n_{sup}$ varies as before between 10k and 8M. Here, unlike in the data-starved scenario of Figure~\ref{fig:experiment-6-8-fixed-small}(a), the benefit from the additional training data is more limited, with error rates actually increasing when $n_{sup}$ is significantly larger than $n_{in}$. This suggests that as training size increases, the skew in train vs.\ test distribution resulting from expansion of the train set quickly outweighs any benefits from the additional training examples. Notably, by the time that $n_{sup}$ approaches 100x the value of $n_{in}$,
error rates have already increased to levels comparable to those at which they are seen to plateau in Figure~\ref{fig:experiment-6-8-fixed-small}(a) for the same choice of $V$,
despite the order of magnitude difference in the amount of in-domain training data.

Results for other choices of $n_{in}$ ranging from 10k up through 500k examples are presented in Appendix~\ref{app:equal-weighting-details}.

\subsection{Over-weighting of Examples from Target Domain}
\label{sect:over-weighting}

\emph{If sample weighting is controlled, then additional data from a related domain can be made to yield a strictly positive effect. However, the performance benefits achievable through this method are limited.}\\

If we had the luxury of being able to train a separate model for each domain, we could consider mitigating the harm caused by the skew in train-test distribution seen in Section~\ref{sect:equal-weighting} by giving special treatment to the in-domain train examples. Here we evaluate the effect of one form of such special treatment by increasing the sampling weight of examples from $W_{TRAIN}$ relative to those from $V_{TRAIN}$ to ensure that at least half of all training observations correspond to in-domain examples.
The target ratio of one-half
matches the optimal ratio observed by \citet{wang2017instance} for a similar instance weighting technique.

Specifically, if, for example, $n_{in}$ is 10k and $n_{sup}$ is 100k, we would duplicate each example of $W_{TRAIN}$ 10 times (prior to shuffling), so that when randomly sampling from the adjusted train set, exactly half of the samples are from $W_{TRAIN}$ and half are from $V_{TRAIN}$. If $n_{sup}$ is less than $n_{in}$, then no adjustment is made.





\begin{figure*}[tb]
  \centering \footnotesize
  (a)\includegraphics[width=0.47\textwidth]{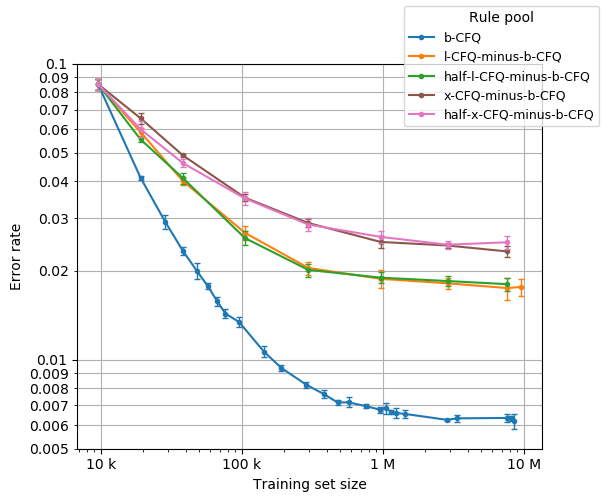}
  (b)\includegraphics[width=0.47\textwidth]{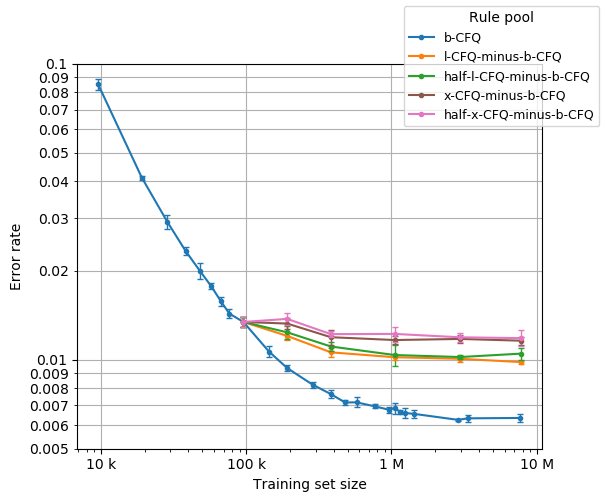}
  \caption{Error rate vs train set size, log-log scale, fixed ratio: \textbf{10k} \bcfq{} examples (a) and \textbf{100k} \bcfq{} examples (b), \bcfq{} test set, \bcfq{} + others train set.
  }
  \label{fig:experiment-6-8-ratio}
  \label{fig:experiment-6-8-ratio-small} 
  \label{fig:experiment-6-8-ratio-large} 
\end{figure*}

Figures~\ref{fig:experiment-6-8-ratio}(a) and (b) again show the results when $n_{in}$ is fixed at 10k and 100k examples, respectively.

As the figures show, if sample weighting is controlled, then additional data from a related domain can be made in most cases to yield a strictly positive effect.

However, again, the more distant the related domain is from the target domain, the more slowly the error rates drop -- and most significantly, the higher the asymptotic value at which they plateau. While we expect that error rates could be further improved by optimizing the exact sampling ratio between $W_{TRAIN}$ and $V_{TRAIN}$,
the results so far suggest that there are limits to the performance benefits achievable through this method.


\section{Related Work}
\label{sect:background}

The relationship between training size and error rates or generalization loss has been studied in a variety of settings. Early research observed that accuracy varies roughly with the log of the training size, both in shallow ML approaches on a natural language disambiguation task~\citep{banko2001scaling} and in convolutional neural networks on an object detection task~\citep{sun2017revisiting}. Others, noting the connection with learning curves observed during training on a large train set, have tried fitting a power law learning curve to the relationship between training size and error rate~\citep{figueroa2012predicting}. 
More recent research has observed consistent power law relationships between training size and cross-entropy error across a range of deep learning architectures, including LSTMs on language modeling and machine translation tasks~\citep{hestness2017deep} and LSTMs or Transformers on language modeling tasks~\citep{rosenfeld2019constructive, kaplan2020scaling, brown2020language},
provided that training size and model size are increased in tandem. In particular \citet{kaplan2020scaling} observe that a highly consistent power law relationship between cross-entropy loss and either of training size or model size persists across many orders of magnitude when not bottlenecked by the other of the two.
Our approach differs from this previous research by specifically investigating the effect of the compositional structure of a task on the training size to error rate relationship.

The degree to which additional training data from a related domain can help or hurt performance has been investigated also in the context of multi-domain learning~\citep{yang2014unified,herzig2017neural,britz2017effective,tars2018multi,mghabbar2020building,wang2020learning} and domain adaptation~\citep{wang2017instance,chu2018survey,zhang2019curriculum,wilson2020survey}.

Our experiments in Section~\ref{sect:equal-weighting} resemble the scenario of multi-domain learning (MDL) in that we aim to train a single model that can apply to multiple domains, not just the single domain from which our test set is drawn. Multi-domain learning seeks, however, to improve performance on individual domains by explicitly distinguishing the domain of each example at train and test time, which is not our focus here. We expect that explicitly distinguishing domains would be less beneficial in the \starcfq{} tasks than in scenarios where the expected output for a given input can differ depending on the domain (something that does not occur in \starcfq{}), but we would welcome investigation into the effectiveness of MDL techniques on \starcfq{} as future work.

Our experiments in Section~\ref{sect:over-weighting} can be considered a form of domain adaptation (DA) in that a model targeting a single domain is trained using a combination of in-domain and out-of-domain data. Our approach corresponds most closely to the \emph{instance weighting} approach, particularly the \emph{batch weighting} variant, which \citet{wang2017instance} found to yield superior performance over other instance weighting techniques. Another popular DA technique which we have not evaluated in our experiments is that of pre-training on a broader data set followed by fine-tuning on in-domain data~\citep{luong2015stanford}, which may be augmented with selections from the out-of-domain data ranked by similarity to the in-domain distribution~\citep{zhang2019curriculum}.

In particular, much attention has been paid recently to the significant improvements in performance achievable on a variety of downstream natural language tasks through pre-training of large scale language models~\citep{devlin2019bert,raffel2019exploring,yang2019xlnet,brown2020language}, including in scenarios of domain adaptation and multi-domain learning~\citep{talmor2019multiqa,gururangan2020don}. 
\citet{talmor2019multiqa} observe that the performance of BERT-large~\citep{devlin2019bert} can be improved on several reading comprehension (RC) benchmarks by additionally pre-training it on a selection of 75k supervised examples from each of 5 different large RC datasets, prior to finally fine-tuning it on the specific target dataset. In cases where the target dataset is large, however, they find that the additional pre-training from the other RC datasets improves performance only about half the time. 
\citet{gururangan2020don} show that applying additional pre-training on unlabeled examples from the target domain improves performance on a number of classification tasks compared to use of RoBERTa~\citep{liu2019roberta} alone.

Our approach differs from this previous research again by
using knowledge of the compositional structure of the task to characterize the relationship between the target domain and related domains, and illustrating the effect of these factors on the slope and limit value of the performance curves.

\section{Conclusion and Outlook}
\label{sect:conc}

In this paper we present \starcfq{}, a suite of large-scale datasets designed for principled investigation into the scalability of ML systems on a compositional NLU task. To the best of our knowledge, \starcfq{} is uniquely suited for this investigation, because it is the first dataset to achieve this scale while providing full details on the compositional structure of each example. We use this dataset to perform experiments illustrating that scalability is indeed a concern even at a scope that is only a tiny fraction of full natural language.

The experiments presented in this paper, however, only scratch the surface of the types of controlled investigations of ML scaling behavior possible using 
\starcfq{}. We hope that this dataset suite will aid the deep learning and NLU communities in the development of more robust and scalable solutions to language understanding.


In particular, we hope to re-use \starcfq{} to evaluate the degree to which language model pre-training and increase of model size affect the scalability curves in the compositional setting. We are interested in exploring the combined effects of compound divergence and rule scope. Further, we are interested to estimate the necessary data sizes and computational power that would be required by current ML approaches to achieve robust levels of language understanding across arbitrary domains and syntax, based on the above results together with an estimate of the effective number of language features in full natural language compared to \xcfq{}.


\section*{Acknowledgements}
We thank Daniel Furrer for code reviews, helpful input in weekly syncs, improving the efficiency of the compound divergence splitting algorithm, and help in running or debugging ML experiments; Xiao Wang for contributing to the planning of grammar extensions; Olivier Bousquet for providing guidance and insights; and Daniel Keysers
for helpful feedback on the paper.

\section*{Ethics Statement}

The purpose of this paper is well aligned with sustainability and efficient use of resources. More scalable models need fewer training examples and thus fewer resources to reach the same accuracy than less scalable ones.

Since the \starcfq{} datasets are generated and ungrounded (except for \ocfq{}), they do not contain any personal data. \ocfq{} does not contain private data as it is grounded on Freebase which is publicly available.

In general, when preparing datasets for use in training language models, care should be taken to avoid inclusion of potentially harmful biases which may be propagated to the language models. Due to their systematic, rule-based generation method, the risk of unintended bias in \starcfq{}, as in \cfq{}, is low compared to datasets mined from large corpora of text in the wild. The main source of potential bias would be in the distribution of entities in the Freebase dataset itself, which may be propagated to the resulting generated questions, particularly during the grounding step. The risk of such bias is reduced in \starcfq{} through the use of ungrounded questions in most of its datasets. In the choice of adjectives modeled in the dataset, some bias (or loss of comprehensiveness) is also introduced due to our choice of only those adjectives whose corresponding entities are relatively frequently used in Freebase.

\bibliography{scalability}

\clearpage
\appendix
\section*{Appendix}

\section{Contributions}
\label{app:contributions}
Dmitry designed most of the experiments and the dataset lattice, generated dataset pools, conducted divergence-related experiments, wrote initial paper draft, contributed to the final paper writing, and made all figures for the paper.

Tibor extended the CFQ dataset with new leaf rules;
wrote the infrastructure for dataset blending and generating reproducible random splits;
conducted all experiments involving random splits and to find an optimal number of training steps;
contributed to writing several sections in the Appendix and the Ethics statement; analyzed the similarity of domains; did the data quality analysis of \xcfq{}; and populated most of the tables used in the paper.

Nathan proposed the paper messaging, researched related work, and wrote the bulk of the main body text.

Nikola supervised the work on extending the CFQ dataset, provided overall technical guidance and contributed to writing Appendix~\ref{app:ungrounded}.

Danila extended the CFQ dataset with the new grammar features, improved scalability of the dataset generation pipeline, did the data quality analysis of \ucfq{}, wrote the corresponding section in the Appendix.

Nathanael provided high-level project direction, wrote the CFQ benchmark section, and provided suggestions while iterating on the messaging.

\section{Grounded vs.\ Ungrounded Datasets}
\label{app:ungrounded}

The \cfq{} dataset was generated using grounding into Freebase. The process of grounding has a couple of effects on the \cfq{} dataset: 1.) It replaces entity placeholders with concrete entities and 2.) it filters out questions which can not be grounded as they are deemed "unnatural".
Unfortunately the grounding process poses a significant challenge in scaling the dataset to a larger number of questions. Grounding is one of the main performance bottlenecks of the \cfq{} generation algorithm due to the limited throughput of the Freebase \SPARQL{} server. In addition, the requirement that every question needs to be grounded in Freebase results in filtering out a large number of natural questions which describe scenarios not present in Freebase. E.g. the question `What Dutch film editor influenced a person.` would be filtered out during grounding as Freebase does not contain a set of entities that satisfy it.

In this work we overcome this problem by skipping the grounding step and hence removing the dependency on the \SPARQL{} server. This allows us to scale the dataset to significantly larger size but it introduces a couple of potentially undesirable effects:

\begin{itemize}
\item \textbf{Over-generation:} it increases the fraction of unnatural questions, many of which could be filtered out based on missing Freebase mappings.
\item \textbf{Skew in the sample distribution:} Grounding in Freebase implicitly applies a semantic filtering of the sentences. Removing the semantic filtering could result in a significant distribution skew on the set of examples which could affect the behavior of Transformers on tasks built on top of this dataset.
\end{itemize}

Given our focus on investigating the scalability of ML systems on the realistic language understanding task represented by the original \cfq{} dataset, our primary concern here is whether the above two effects could lead to materially different scaling behaviors of ML systems on the ungrounded datasets compared to the original grounded one.

In order to address this concern we perform a series of experiments aimed at determining the effect of grounding on the performance of Transformers. Based on the results of those experiments we conclude that the scalability experiments performed in this paper over the ungrounded \starcfq{} dataset suite should be indicative for the behavior of Transformers on grounded datasets.

In section Appendices~\ref{app:ocfq} and~\ref{app:ucfq-quality} we describe the pair of datasets used to perform the aforementioned experiments and in Appendices~\ref{app:exp-3.1} and~\ref{app:exp-1.1} we describe the experiments in detail.

\subsection{\ocfq{} and \ucfq{}}
\label{app:ocfq}

In order to facilitate a set of experiments which evaluate the effect of grounding we generated a pair of datasets (\ocfq{} and \ucfq{}) which only differ by the fact that \ocfq{} is grounded, while \ucfq{} is not.

\ocfq{} and \ucfq{} are generated from the same set of rules, which we call the \ucfq{} rule set. The \ucfq{} rule set is very close to the set of rules used in the original \cfq{} dataset (see Appendix~\ref{app:ucfq-rule-set} for minor differences). This set of rules was used in two different ways in order to generate the two datasets:

\begin{itemize}

\item \textbf{\ocfq{}}: The initial set of examples were generated by using the \ucfq{} rules to parse the questions from the \cfq{} dataset, yielding the same questions as in \cfq{}, but instrumented with atoms and compounds based on the \ucfq{} grammar, and with potentially different SPARQL output due to slightly adjusted property mappings (see Appendix~\ref{app:ucfq-rule-set}). Due to the minor differences between the \cfq{} and \ucfq{} grammars, a small fraction (4.4\%) of the \cfq{} questions were not parsable by the \ucfq{} grammar. These questions were discarded.

\item \textbf{\ucfq{}}: \ucfq{} was produced using the randomised generation procedure described in \citet{keysers2020}, except that the grounding step was skipped.
\end{itemize}

For more details on the \ucfq{} rule set, see Appendix~\ref{app:ucfq-rule-set}.

\subsection{Data Quality Analysis of \ucfq{}}
\label{app:ucfq-quality}

During the development of our data generation pipeline, we manually checked the generated examples for quality. Below is a random selection of 50 examples of the final \ucfq{} dataset (no cherrypicking was used). 

\begin{enumerate}
    \item What did a costume designer that M1 played and M2 played found?
    \item What male actor was a cinematographer and editor of M2?
    \item Was M5 a Japanese Dutch male Spanish screenwriter's country of nationality?
    \item What was directed and executive produced by a Swedish male costume designer that a actor played?
    \item What was edited and directed by a German Japanese French male art director's parent's sibling?
    \item What Chinese employee and founder of M1 influenced M0?
    \item Was M2 played by a male cinematographer and star of M0 and played by M3?
    \item Which female writer and executive producer of M0 and M1 was a Canadian Swedish Chinese editor of M6?
    \item Who was played by a film editor's parent's parent and spouse?
    \item Was M1's cinematographer and producer a Dutch female art director's country of nationality's Swedish male art director?\label{debatable_ucfq_2_1}
    \item Was M3 founded by M1's female Japanese parent?
    \item Was M4's child a Canadian female Spanish Italian costume designer of M0?
    \item Was M2's star M1's Spanish male child?
    \item Who was a male child of a Japanese film editor?
    \item Was M1 written by M0's director, writer, editor, and art director?
    \item Which French parent of a costume designer did a male costume designer marry?
    \item What film was written by, edited by, directed by, and executive produced by a character's Italian spouse?
    \item Was M0's writer, cinematographer, director, art director, costume designer, editor, executive producer, and producer a film's producer?
    \item Who was M3's producer's female Dutch employee?
    \item Did M1 marry and influence a art director's sibling, child, and spouse?
    \item What was produced and distributed by a costume designer's female Chinese Swedish Japanese parent?
    \item Who married M3's Italian female Chinese parent?
    \item Was a male Spanish child of a Japanese art director's child M2's Chinese star?
    \item Did M1 direct M2 and write a prequel of M0?
    \item What did a Italian male writer of M0, M1, M2, and M3 produce?
    \item Who was employed by M2's parent, was influenced by a film producer, and was employed by M1?
    \item Did a Japanese executive producer, art director, producer, and star of M0, M1, and M2 executive produce M3?
    \item Was M1's Mexican spouse a female German screenwriter's parent, sibling, spouse, and child?\label{debatable_ucfq_1_1}
    \item Did M1 influence a art director, costume designer, and cinematographer of M0, write M2, and marry M3?
    \item Was a film editor a male art director that married M3?
    \item Was M5 executive produced by a female Spanish American Swedish art director's parent and child?\label{debatable_ucfq_1_2}
    \item Who was a Dutch Swedish costume designer's Swedish Mexican sibling?
    \item Which male spouse of M2's child was played by M0?
    \item Which writer, cinematographer, director, producer, executive producer, star, and costume designer of M0 produced M1's sequel?
    \item Was M0 a producer, writer, executive producer, cinematographer, costume designer, and star of M1, M2, M3, M4, and M5?
    \item Who influenced M2's female editor?
    \item Was a film's star M0's producer, costume designer, cinematographer, director, executive producer, writer, art director, and editor?
    \item Was a parent, child, sibling, and spouse of a art director a female child of M0?\label{debatable_ucfq_1_3}
    \item What film was executive produced by a cinematographer and starred M1?
    \item Who was a male actor that married and influenced a female costume designer's male child?
    \item Did a female Swedish British film director's child acquire M0?\label{debatable_ucfq_2_2}
    \item Which Chinese Mexican founder of M0 produced M3?
    \item Which Spanish Italian film director was played by M1?
    \item What did a French employee of a film producer write?
    \item Was a person M1's art director's parent's parent?
    \item Were M2, M3, M4, M5, M6, M7, and M8 edited by M1's male editor?
    \item Who was a American child of a male American German French actor's American parent?
    \item Was a child and parent of a company M0's star and executive producer?\label{debatable_ucfq_2_3}
    \item Was M4's female Canadian Italian French Chinese child's gender M6?
    \item Did a male Canadian person whose spouse founded M4 edit M1?
\end{enumerate}

Manual checking showed that all SPARQL queries associated with these questions are semantically correct, but several examples sound unnatural. These examples can be divided in two groups:

\begin{enumerate}
    \item Questions that require an entity to fill multiple roles of another entity which for some pairs of roles is impossible in reality (e.g. someone's parent and child cannot be the same person). The 3 examples in this group are \ref{debatable_ucfq_1_1}, \ref{debatable_ucfq_1_2}, \ref{debatable_ucfq_1_3}. 
    \item Questions, where an entity is used with a property that does not match the type of that entity (e.g. ''country of nationality's art director''). There are 3 such questions in our sample: \ref{debatable_ucfq_2_1}, \ref{debatable_ucfq_2_2}, \ref{debatable_ucfq_2_3}.
\end{enumerate}

Comparing these results with quality analysis of the \cfq{} dataset described in \citet{keysers2020}, we see that the unnatural-sounding questions are approximately two times more frequent in \ucfq{} that can be attributed to the fact that grounding in Freebase performed for \cfq{} allowed to eliminate at least some questions with incompatible roles and/or properties.

The presence of these questions in the \ucfq{} dataset doesn't materially affect any of our findings, because every such question can still be unambiguously represented as a \SPARQL{} query.

\subsection{Effect of Testing on Ungrounded vs.\ Grounded Data}
\label{app:exp-3.1}

To explore the degree to which ML systems' scalability behaviors differ between \ucfq{} and \ocfq{}, we generate random train/test splits with varying train set sizes from each of these datasets and compare the accuracy trajectories of the trained models.



The training sizes for \ucfq{} range from 9.5k up to 8.7M examples. The training sizes for \ocfq{} range from 9.5k up to 640k examples (larger sizes omitted for \ocfq{} due to the limited size of the grounded dataset).

\begin{figure}[tb]
   \centering
   \includegraphics[width=0.47\textwidth]{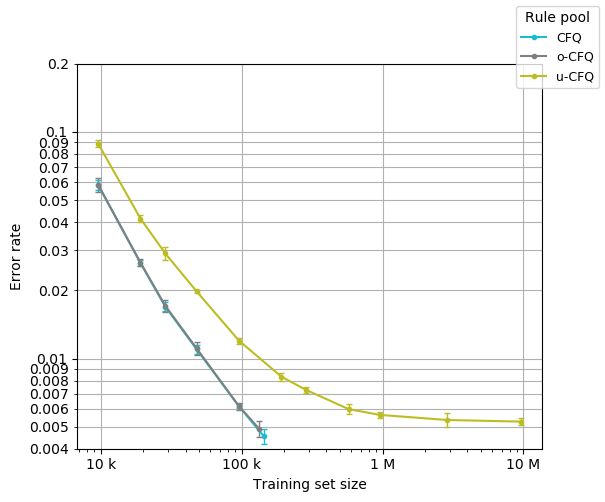}
   \caption{Effect of grounding on scalability curve. Accuracy vs train set size for random splits of \cfq{}, \ocfq{}, and \ucfq{}. Note that the curves for \cfq{} and \ocfq{} are virtually identical and largely overlap.}
   \label{fig:experiment-3.1}
\end{figure}

The results are presented in Figure~\ref{fig:experiment-3.1}.




We observe that while error rates on \ucfq{} are somewhat higher than on \ocfq{}, error rates follow qualitatively similar trajectories on both sets.

Based on this, we conclude that performing experiments on ungrounded data should be a reasonable proxy for experimenting on grounded data and that we can expect findings from experiments on ungrounded \cfq{}-like datasets to carry over to the case of grounded datasets as well. For the bulk of this paper, we thus focus primarily on experiments on ungrounded data due to the increased feasibility of generating large quantities of ungrounded data.

\subsection{Effect of Training on Ungrounded vs.\ Grounded data}
\label{app:exp-1.1}




\begin{figure}[tb]
   \centering
   \includegraphics[width=0.47\textwidth]{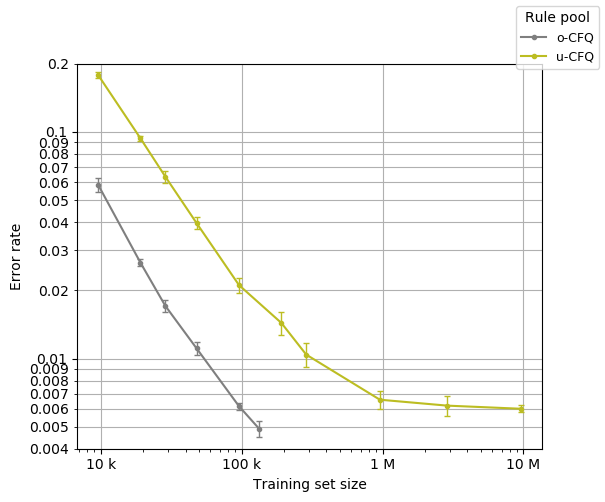}
   \caption{Efficacy of training on ungrounded data when targeting grounded data as the test set. Accuracy vs train set size when \ocfq{} is used as the test set, and either \ocfq{} or \ucfq{} as the train set. While at any given training size, training on ungrounded data leads to lower accuracies than when training on grounded data from the same distribution as the test set, if significantly larger amounts of ungrounded data are available, then the larger training size can compensate for this accuracy loss.}
   \label{fig:experiment-1.1}
\end{figure}


While the \cfq{} and \starcfq{} datasets use rule-based data generation primarily as a means of constructing a controlled testing environment for understanding the behavior of ML systems, a similar auto-generation approach, if applied across a broader range of domains, could conceivably be used as a means to provide or augment training data for a system intended to be applied in real-life scenarios. One question that arises in such a scenario is how effective it would be to apply the cheaper process of generating (potentially large numbers of) ungrounded training examples, compared to generating a smaller number of more realistic training examples via the more expensive grounding process.

To explore this trade-off between data size and data purity, we conduct an experiment in which the test set consists of ``realistic'' grounded examples from \ocfq{}, while the train set consists of ungrounded examples from \ucfq{}, with the train size ranging from 9.5k up to 8.7M examples. The resulting accuracies are compared with a baseline of a pure random split on \ocfq{}.

The results are presented in Figure~\ref{fig:experiment-1.1}.

As expected, at any given train size, training on ungrounded data leads to lower accuracy than training on grounded data from the same distribution as the test set. Accuracy increases to competitive levels, however, as the amount of ungrounded training data increases, suggesting that if an auto-generation process were able to generate significantly larger amounts of ungrounded training data than could be acquired for grounded examples, the larger training size can compensate for the nominal accuracy loss that results from the inclusion of ``unrealistic'' examples in the train set.


\section{Dataset Stats}
\label{app:datasets}

Table~\ref{tab:datasets} shows a complete list of the datasets included in \starcfq{}, with characteristics including their size and complexity in comparison with the original \cfq{} dataset.
    
\begin{table*}[tb]
    \begin{tabular}{@{}llrrrr@{}}
        \hline
        {} & {} & {} & {}                   & Question & Query \\
        Dataset & Description & \hspace{-2ex}Questions & Queries & patterns & patterns \\
        \hline
        \hline
        \cfq{} & From \citet{keysers2020} & 239,357 & 228,149 & 49,320 & 34,921 \\
        \hline
        \ocfq{} & Slightly filtered and re-instrumented & 228,901 & 165,736 & 44,465 & 34,278 \\
        \ucfq{} & Larger ungrounded version & 9,925,221 & 6,551,678 & 319,407 & 762,680 \\
        \hline
        \bcfq{} & Base of rule set lattice & 8,663,352 & 4,617,765 & 419,611 & 696,840 \\
        \ncfq{} & Base + additional non-leaf rules & 8,716,725 & 4,580,018 & 608,651 & 841,749 \\
        half-\ncfq{} & Base + half the additional non-leaf rules & 8,699,762 & 4,385,655 & 452,819 & 706,452 \\
        \lcfq{} & Base + additional leaf rules & 9,874,258 & 7,266,600 & 618,079 & 1,630,225 \\
        half-\lcfq{} & Base + half the additional leaf rules & 9,297,045 & 6,656,162 & 558,990 & 1,484,723 \\
        \xcfq{} & Base + both types of additional rules & 9,879,894 & 7,249,705 & 713,137 & 1,847,555 \\
        half-\xcfq{} & Base + half of both types of rules & 9,294,958 & 6,469,571 & 530,405 & 1,535,828 \\
        \hline
        \hline
    \end{tabular}
    \caption{A list of all datasets included in \starcfq{}, with their characteristics. The first row is the original \cfq{} from \citet{keysers2020}, for comparison. The second section are newly introduced datasets based on roughly the same rule set as \cfq{}. The third section are the datasets corresponding to the nodes of the rule set lattice described in Section~\ref{sect:rule-set-lattice}. 
    }
    \label{tab:datasets}
\end{table*}

\section{Rule Sets}
\label{app:rule-sets}

\subsection{\cfq{} / \ucfq{} Rule Set}
\label{app:ucfq-rule-set}

The examples from \cfq{} that are not included in \ucfq{} can be split to the following groups:


\begin{enumerate}
    \item 9037 questions mentioning actor as a role, for example ``Did M1 influence M0's actor?''. This includes unnatural questions such as ``Was M1 a company's actor?'', and moreover, it was always assumed that it referred to an actor of a character (and not an actor of a movie) which sometimes resulted in non-obvious \SPARQL{} queries. In lieu of a better solution, all questions containing actor as a role have been removed.
    \item 1306 unnatural questions that treat countries of nationalities as organizations, for example ``Who did M1's country of nationality employ?''.
    \item 930 questions that require a non-person writer, for example ``Was a company M1's writer?''.
    \item 34 questions containing the unnatural phrase ``playing a film'', for example ``Who played a film?''.
\end{enumerate}

The generated \SPARQL{} can also be different for the same question in \cfq{} and \ocfq{}, due to the following differences in the \ucfq{} rule set:

\begin{enumerate}
    \item Directing, producing and writing were mapped in \cfq{} solely to movie-related Freebase properties. The \ucfq{} rule set covers also TV programs so these actions are mapped to the disjunction of the movie and TV program related Freebase properties. For example, \ucfq{} maps ``Who directed M0?'' to a \SPARQL{} query representing a person directing a movie or a TV program. In the \starcfq{} datasets, this is important to ensure that the \SPARQL{} is independent of the type of M0. In the case of \ocfq{} and \ucfq{}, this is just a technicality that makes these datasets compatible with the \starcfq{} datasets.
    \item Knowledge rules in the \cfq{} rule set map genders only to the Freebase property representing the gender of a real person. The \ucfq{} rule set maps genders to a disjunction of Freebase properties corresponding to the gender of a real person and the gender of a fictional character. This seems like an oversight in \cfq{} since it already covered fictional characters in other cases.
\end{enumerate}

\subsection{\bcfq{}}
\label{app:bcfq}
A natural choice for the bottom of the rule set lattice would be the \ucfq{} rule set, as it was the starting point of our rule set extension that lead to generation of \xcfq{}. This choice, however, would be sub-optimal with respect to one of the desired criteria for the experiments we plan to run.

Specifically, one of the goals of this research is to compare the effect of rule scope increase from addition of leaf rules vs. addition of non-leaf rules. In order to do this,
we would ideally like to separate the new rules into two disjoint sets: one consisting of only new leaf rules (for inclusion in \lcfq{}) and one consisting of only new non-leaf rules (for inclusion in \ncfq{}).

A clean separation is difficult, however, due to certain interdependencies between the leaf and non-leaf rules. For example,
one of the additions we make for \xcfq{} is to introduce film genres. While it naturally counts as a leaf change (29 added leaf rules), its support requires the implementation of a number of non-leaf rules as well. If \ucfq{} is the bottom of the lattice, this would prevent us from having a clear separation between leaf and non-leaf rules.

In order to mitigate this problem, we define a new rule set \bcfq{} to service as the lattice's bottom. We design it to be as close as possible to \ucfq{} (in terms of size and complexity), but allowing clear separation of leaf and non-leaf rule additions. \bcfq{} differs from \ucfq{} in the following ways:
\begin{itemize}
    \item Adds 2 non-leaf and 2 leaf grammar rules related to genres.
    \item Removes all rules related to gender.
    \item Removes a leaf rule corresponding to ``what'' in certain places.
    \item Removes 10 leaf rules corresponding to nationalities (keeping the other 2 rules).
\end{itemize}

These changes aim to maintain a balance between added/removed rules while maintaining the additivity property of the rule set lattice.

\subsection{Rule Set Lattice}
\label{app:lattice}

\begin{figure}[t!]
   \centering
   \includegraphics[width=0.47\textwidth]{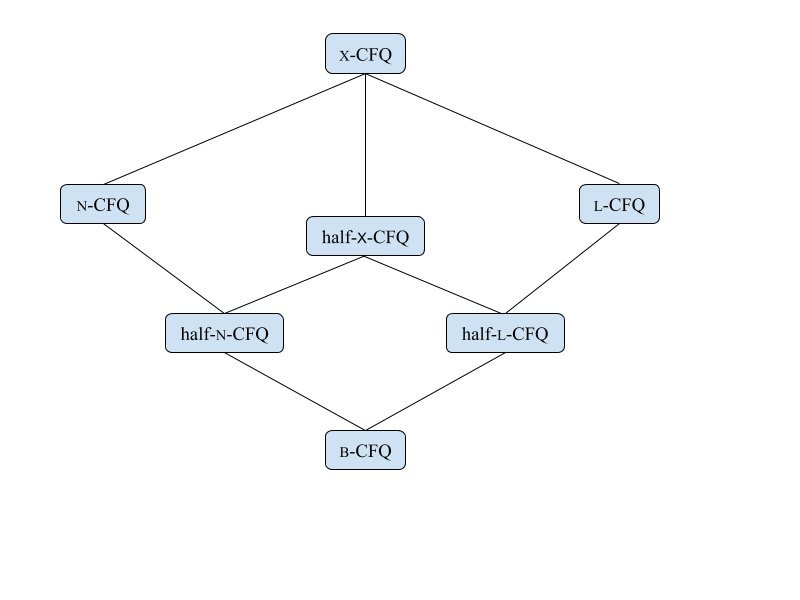}
   \caption{Rule set lattice, where edges correspond to an inclusion relation. \bcfq{} contains the rules shared by all other rule sets in the lattice. \ncfq{} adds more non-leaf rules. \lcfq{} adds more leaf rules. \xcfq{} contains the union of both.}
   \label{fig:lattice}
\end{figure}

Rule set lattice, corresponding to \starcfq{}, is illustrated in Figure~\ref{fig:lattice}, in which each rule set is a superset of its child rule set(s).

The description of rule sets \lcfq{}, \ncfq{} follows directly from the rule sets \xcfq{} and \bcfq{}, taking into account the additivity principle described in Appendix~\ref{app:bcfq}. There are more choices for the half datasets available. Below we describe the reasons behind the particular rule set's content.  
\begin{description}
\item[half-\lcfq{}.] For this rule set we go through all newly supported Freebase types and properties (see Table~\ref{tab:grammar-features-full} for details and examples) and pick half of the leaves in each categories, rounded down. We also include all the surface forms in this rule set.
\item[half-\ncfq{}.] It is harder to choose the rules for this set, as individual rules are not independent: some of them are required for the others to be able to trigger. In the end we separate the rules of this kind into two roughly equal groups: the preposition-related rules and all other rules. This rule set contains all preposition-related rules.
\item[half-\xcfq{}.] Similar to \xcfq{}, this rule set is a union of half-\lcfq{} and half-\ncfq{}.
\end{description}

Table~\ref{tab:rules-for-grammar-features} shows statistics about the newly supported surface forms and the newly added Freebase type, property and entity mappings.

\begin{table*}[tb]
    \begin{tabular}{@{}lrrrrr@{}}
        \hline
        & & & \# added & \# added & \# added \\
        & \# affected & \# affected & Freebase & Freebase & Freebase \\
        New language feature & grammar rules & tokens & properties & types & entities \\
        \hline
        \hline
        TV program and contributions & 32 & 20 & 10 & 5 & 0 \\
        Film genre & 30 & 16 & 2 & 0 & 13 \\
        Religion & 27 & 12 & 2 & 1 & 8 \\
        Ethnicity & 18 & 9 & 2 & 1 & 6 \\
        Movie language & 16 & 11 & 2 & 1 & 9 \\
        Additional family relations & 16 & 8 & 0 & 0 & 0 \\
        Additional film related verbs & 9 & 8 & 8 & 1 & 0 \\
        Generic notion of ``contributing to'' a film & 4 & 2 & 2 & 0 & 0 \\
        Woman/Man & 4 & 2 & 0 & 0 & 0 \\
        Celebrity & 2 & 1 & 0 & 1 & 0 \\
        Friend & 1 & 1 & 2 & 0 & 0 \\
        \hline
        `Script(-)writer' (meaning `Screenwriter') & 2 & 2 & 0 & 0 & 0 \\
        `Offspring' (meaning `Child' of a person) & 2 & 1 & 2 & 0 & 0 \\
        `Subsidiary' (meaning `Child' of a company) & 2 & 1 & 2 & 0 & 0 \\
        `Movie' (meaning `Film') & 1 & 1 & 0 & 0 & 0 \\
        `Which' (synonym for `That') & 1 & 1 & 0 & 0 & 0 \\
        Definite article & 1 & 1 & 0 & 0 & 0 \\
        \hline
    \end{tabular}
    \centering
    \caption{Statistics for newly supported surface forms, Freebase types, properties and entities.}
    \label{tab:rules-for-grammar-features}
\end{table*}


\begin{table*}[tb]
    \begin{tabular}{@{}ll@{}}
        \hline
        New language feature & Example question \\
        \hline
        \hline
        Verb + preposition (with object) & Who \featurespan{played} M3 \featurespan{in} M2 \\
        Verb + preposition (without object) & \featurespan{In} which TV program did M1 \featurespan{play} \\
        AND of prepositional phrases & Who played \featurespan{in M1 and in M2} \\
        AND of determined noun phrases & Did M1's father marry \featurespan{M2 and a TV program creator} \\
        AND of possessives & Which \featurespan{Mexican German adherent of M5 and sister of M0} created M3 \\
        AND of verb phrases (with object) & Did M0 \featurespan{direct M2 and marry M1} \\
        Additional passive sentence types & \featurespan{Which} company \featurespan{was} M1 employed \featurespan{by} \\
        \hline
        TV programs (and related properties) & What did a \featurespan{creator} of M1's \featurespan{TV program} executive produce \\
        Film genre & Who was a \featurespan{crime fiction film}'s Indian writer \\
        Religion & What did a \featurespan{religion}'s Mexican \featurespan{adherent} found \\
        Ethnicity & What was M1's Anglican art director's \featurespan{ethnicity} \\
        Movie language & Did M2 distribute a \featurespan{French speaking} film \\
        Additional family relations & Was the \featurespan{daughter} of the \featurespan{brother} of M0 M2 \\
        Additional film-related verbs & Did a scriptwriter's mother \featurespan{art direct} M1 \\
        Generic notion of ``contributing to'' a film & Did M2's husband \featurespan{contribute to} M0 \\
        Woman/Man & Did M1's husband play a \featurespan{woman} \\
        Celebrity & Was the art director of M1 a \featurespan{celebrity}'s father \\
        Friend & Who was a \featurespan{friend} of a TV producer's child \\
        \hline
        `Script(-)writer' (meaning `Screenwriter') & Was a \featurespan{script-writer} M1's Canadian employee \\
        `Offspring' (meaning `Child' of a person) & Who was M3's \featurespan{offspring}'s African-American son \\
        `Subsidiary' (meaning `Child' of a company) & What was M0's distributor's parent's \featurespan{subsidiary} \\
        `Movie' (meaning `Film') & Was M0 a German speaking family \featurespan{movie} \\
        `Which' (synonym for `That') & Was M0 a film director \featurespan{which} M2 married \\
        Definite article & Did \featurespan{the} spouse of a film editor create M0 \\
        \hline
    \end{tabular}
    \centering
    \caption{Examples of questions enabled via newly added grammar rules. The first section represent newly supported grammatical constructs. The middle section represent newly supported Freebase types and properties. The last section represent addition of new surface forms that are treated as synonyms for an existing one.}
    \label{tab:grammar-features-full}
\end{table*}


\subsection{Data Quality Analysis of \xcfq{}}

We also analysed the data quality of \xcfq{}. Below is a random selection of 50 examples of the final \xcfq{} dataset (again, no cherrypicking was used).

\begin{enumerate}
    \item What executive producer and producer of M0 was M1's editor, art director, star, costume designer, and executive producer?
    \item Was M5 a art director whose Mormon spouse, father, husband, and wife produced M0 and M1?
    \item Was a Hindu Jewish Chinese Irish American brother of a drama film's producer's child M2?
    \item Was M2 a daughter, spouse, and sibling of the thriller movie's executive producer, costume designer, director, cinematographer, and star?\label{debatable_xcfq_1_1}
    \item Was M2's child's friend, offspring, parent, wife, and sibling M0's child and parent?\label{debatable_xcfq_1_2}
    \item Who was M2's Hindu costume designer's husband, sibling, friend, and parent?\label{debatable_xcfq_1_3}
    \item Was a movie whose art director and executive producer influenced and was influenced by a film distributor's employee and founder M1?
    \item Who was the Dutch sibling and husband of M0?\label{debatable_xcfq_1_4}
    \item Which Buddhist Mormon male Muslim parent of M6's star was M1's Hindu spouse?
    \item Who was a parent of a company's white Catholic Canadian founder?
    \item What female Irish American Indian child of a cinematographer did M2 and M3 employ?
    \item Was a Hindu costume designer's parent, spouse, and son a Mexican child of the creator and producer of M2?\label{debatable_xcfq_1_5}
    \item Was M3 a actor whose parent and wife distributed and produced M0 and M1?
    \item Was M2's Hindu spouse a English wife of M3?
    \item Was M2's producer's Buddhist sibling a Jewish TV director's daughter?
    \item Was the woman's spouse, sibling, father, and child M2?\label{debatable_xcfq_1_6}
    \item Who was the child, sibling, parent, and husband of M2's friend and sister?\label{debatable_xcfq_1_7}
    \item Was M0's art director, costume designer, executive producer, producer, writer, editor, and star M2's mother?
    \item Was the man the executive producer of M1, M2, M3, M4, and M5?
    \item Was M2 a drama sequel and prequel of the German speaking film?\label{debatable_xcfq_1_8}
    \item Which husband of the person and atheist art director of M1 and M2 were M4, M5, and M6 directed by?
    \item Was the TV show's art director M1's mother?
    \item What did the production company's Canadian white child's French mother's friend distribute?\label{debatable_xcfq_2_1}
    \item What was M0's Buddhist Anglican producer's employer's parent and subsidiary?\label{debatable_xcfq_1_9}
    \item What was a country of nationality of M5's American female Italian Irish American Japanese spouse?
    \item Was a screenwriter M1's friend, offspring, and mother?\label{debatable_xcfq_1_10}
    \item Was M1's Dutch producer the art director's child, wife, and friend?\label{debatable_xcfq_1_11}
    \item Was the religion's Catholic white costume designer M4's Muslim executive producer's spouse, mother, and friend?\label{debatable_xcfq_1_12}\label{debatable_xcfq_2_2}
    \item Was a German parent of M1 a father and spouse of M3's adherent?\label{debatable_xcfq_1_13}
    \item Was M1 a costume designer's Mormon English Buddhist Indian white wife?
    \item Who was a African-American son of M3's sibling?
    \item Did a Hindu Irish American person's Hindu Muslim mother write M0?
    \item Were M2, M3, M4, M5, M6, M7, M8, and M9 created by M1's cinematographer and M0's child?
    \item Was a costume designer's husband's daughter's husband's race M3?
    \item Was M2 M1's female producer's sister's sibling and mother?\label{debatable_xcfq_1_14}
    \item Who was a religion's costume designer's English Chinese Jewish son's brother and child?\label{debatable_xcfq_2_3}
    \item Was M3 the child of the Swedish speaking animation film's sequel's producer, art director, and editor?
    \item Who was the Dutch speaking action movie's star's child, spouse, father, and sibling?\label{debatable_xcfq_1_15}
    \item Who was the parent of a film director's white female Buddhist Jewish French Anglican spouse?
    \item What was produced by a character's parent's founder?\label{debatable_xcfq_2_4}
    \item Who was a Muslim mother of the Muslim woman?
    \item Who was a screenwriter's spouse, brother, child, and parent?\label{debatable_xcfq_1_16}
    \item Was M1 the costume designer's Muslim wife?
    \item What did a Indian Mexican TV producer whose employer wrote M3 create?\label{debatable_xcfq_2_5}
    \item Did the science fiction movie's Hindu English white director write M0 and M1?
    \item Did a brother of a person distribute M0 and M1?\label{debatable_xcfq_2_6}
    \item Who was a Canadian male TV program creator whose child was acquired by and acquired the TV writer's child?\label{debatable_xcfq_1_17}\label{debatable_xcfq_2_7}
    \item Was a TV programme's executive producer's mother, friend, and offspring M1?\label{debatable_xcfq_1_18}
    \item What did a TV director's religion's TV program star?\label{debatable_xcfq_2_8}
    \item Who was M3's Anglican Catholic African American husband?
\end{enumerate}

Manual checking also indicated that all questions are associated with the semantically correct \SPARQL{} queries, but similarly to \ucfq{}, several examples sound unnatural. These examples can again be divided in two groups:

\begin{enumerate}
    \item Questions that require an entity to fill multiple roles of another entity which for some pairs of roles is impossible in reality (e.g. someone's parent and child cannot be the same person).
    The 18 examples in this group are \ref{debatable_xcfq_1_1}, \ref{debatable_xcfq_1_2}, \ref{debatable_xcfq_1_3}, \ref{debatable_xcfq_1_4}, \ref{debatable_xcfq_1_5}, \ref{debatable_xcfq_1_6}, \ref{debatable_xcfq_1_7}, \ref{debatable_xcfq_1_8}, \ref{debatable_xcfq_1_9}, \ref{debatable_xcfq_1_10}, \ref{debatable_xcfq_1_11}, \ref{debatable_xcfq_1_12}, \ref{debatable_xcfq_1_13}, \ref{debatable_xcfq_1_14}, \ref{debatable_xcfq_1_15}, \ref{debatable_xcfq_1_16}, \ref{debatable_xcfq_1_17}, and \ref{debatable_xcfq_1_18}. 
    \item Questions, where an entity is used with a property that does not match the type of that entity (e.g. ``religion's costume designer''). There are 8 such questions in our sample: \ref{debatable_xcfq_2_1}, \ref{debatable_xcfq_2_2}, \ref{debatable_xcfq_2_3}, \ref{debatable_xcfq_2_4}, \ref{debatable_xcfq_2_5}, \ref{debatable_xcfq_2_6}, \ref{debatable_xcfq_2_7}, and \ref{debatable_xcfq_2_8}.
\end{enumerate}

The presence of these questions in the \xcfq{} dataset doesn't materially affect any of our findings either, because every such question can still be unambiguously represented as a \SPARQL{} query.

\section{Hyperparameters and Hardware}
\label{app:hyperparameters}

Unless explicitly stated otherwise, for all experiments in this paper, we use a Transformer architecture with hyperparameters identical to those used in \citet{keysers2020}, with the exception of the training steps, which we increase to $500,\!000$, instead of the original $35,\!000$.

\begin{figure}[tb]
   \centering
   \includegraphics[width=0.47\textwidth]{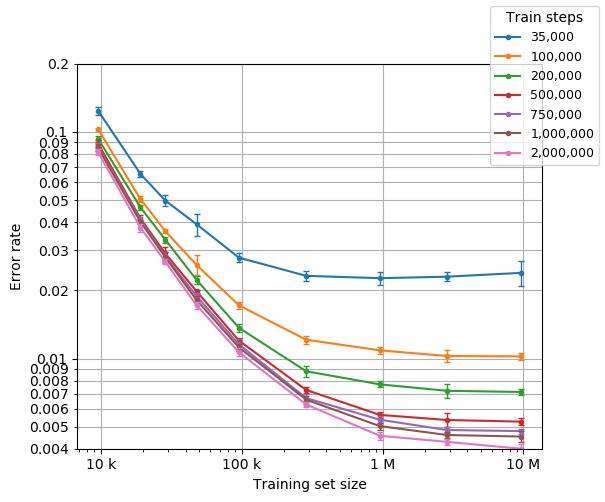}
   \caption{Error rate vs number of training steps on a \ucfq{} random split. Training for more steps reduces error rate consistently, particularly at large training sizes, but with diminishing returns after $500,\!000$ steps. }
   \label{fig:train-steps}
\end{figure}

We measured the impact of changing this metric and found $500,\!000$ to be the best trade-off between computational cost and usefulness, given our choice of model size for the experiments (see Figure~\ref{fig:train-steps}).

The hyperparameters used are summarized in Table~\ref{tab:hyperparameters}.

\begin{table}[tb]
    \begin{tabular}{@{}lr@{}}
    \hline
        Hyperparameter &  Value \\
        \hline \hline 
        train steps             & 500,000 \\
        batch size              & 4,096 \\
        hidden size             & 128 \\
        num hidden layers       & 2 \\
        num heads               & 16 \\
        learning rate schedule  & 0.08*linear\_warmup \\
                                & *rsqrt\_decay \\
        learning rate warmup steps   & 4,000 \\
        \hline
    \end{tabular}
    \centering \small
    \caption{Summary of hyperparameters that deviate from the defaults of the \textit{transformer\_base} hyperparameter set from the tensor2tensor framework \citep{tensor2tensor}.}
    \label{tab:hyperparameters}
\end{table}

The experiments are run on a variety of  hardware depending on the resource requirements of particular experiment stages. All stages that could benefit from ML acceleration (namely, training and evaluation) are run using NVidia P100 GPUs.
The most memory-intensive parts of the experiments (namely, preprocessing of splits) are run on a cluster of Cascadelake 2200MHz servers with 767Gb of memory. 

\section{Domain Size Estimate}
\label{app:domain-size}

\begin{table*}[tb]
    \begin{tabular}{@{}lrrrrrr@{}}
        \hline
        Domain &  & Relative & Vocabulary & Rule & Rule occurrence \\
        (rule set) & Baseline & size & overlap & overlap & overlap \\
        \hline
        \hline
        \ncfq{} & \bcfq{} & $1.28x$ & $98.88\%$ & $90.53\%$ & $98.02\%$ \\
        half-\ncfq{} & \bcfq{} & $1.07x$ & $98.88\%$ & $92.15\%$ & $98.70\%$ \\
        \ncfq{} & half-\ncfq{} & $1.21x$ & $100.00\%$ & $98.24\%$ & $99.31\%$ \\
        \hline
        \lcfq{} & \bcfq{} & $26.14x$ & $54.32\%$ & $55.84\%$ & $79.97\%$ \\
        half-\lcfq{} & \bcfq{} & $9.24x$ & $66.67\%$ & $68.84\%$ & $83.28\%$ \\
        \lcfq{} & half-\lcfq{} & $6.08x$ & $81.37\%$ & $80.89\%$ & $94.02\%$ \\
        \hline
        \xcfq{} & \bcfq{} & $32.45x$ & $52.73\%$ & $51.44\%$ & $77.41\%$ \\
        half-\xcfq{} & \bcfq{} & $10.14x$ & $66.17\%$ & $65.03\%$ & $81.73\%$ \\
        \xcfq{} & half-\xcfq{} & $7.57x$ & $80.12\%$ & $79.10\%$ & $92.97\%$ \\
        \hline
        \lcfq{} & \ocfq{} & $21.50x$ & $56.17\%$ & $60.05\%$ & $83.07\%$ \\
        \xcfq{} & \ocfq{} & $26.07x$ & $55.15\%$ & $55.32\%$ & $81.20\%$ \\
        \hline
    \end{tabular}
    \centering
    \caption{Metrics estimating the relative size and proximity of the domains of the \starcfq{} rule sets.}
    \label{tab:domain-size}
\end{table*}

We use the term \emph{domain} in a sense compatible with that used, for example, by \citet{gururangan2020don}, who defines this as ``a distribution over language characterizing a given topic or genre''. More specifically, given our focus on rule-generated datasets, we characterize domains in terms of sets of grammar rules that determine both the vocabulary and syntax typical of those domains, and focus primarily on the binary distinction of whether a given sentence is ``in domain'' or ``out of domain'', rather than the precise probability of the sentence's occurrence. From this perspective, for the purposes of this paper, we can consider \emph{domain} to refer more simply to the set of all possible sentences generated by a given set of grammar rules -- also referred to in grammar research as the \emph{language} generated by the given grammar~\citep{chomsky1956three}.

Table~\ref{tab:domain-size} summarizes the estimated size of each of the domains in the \starcfq{} rule set lattice relative to relevant baseline domains, along with several other metrics that can be considered indicators of domain proximity. The metrics displayed are defined as follows:
\begin{itemize}
    \item \emph{Relative size} is calculated based on the ratio of examples in the dataset associated with the given domain that are also in the baseline domain. The underlying assumption is that the datasets have a roughly even distribution of examples in the domain that they cover.
    The step in the generation pipeline that subsamples the original, larger pools to maximize rule divergence helps to achieve this.
    \item \emph{Vocabulary overlap} indicates the fraction of tokens used in the given rule set that are also used in the rule set of the baseline domain. This is comparable to the ``vocabulary overlap'' used by \citet{gururangan2020don} as a rough indicator of domain proximity. While \citet{gururangan2020don} calculate this based on the top 10k most frequent words from each domain, in our case we consider all tokens from the domain, given the relatively small number of tokens used.
    \item \emph{Rule overlap} indicates the fraction of rules in the given rule set that are also in the rule set of the baseline domain. 
    \item \emph{Rule occurrence overlap} indicates the fraction of occurrences of rules across all examples in the dataset associated with the given domain that correspond to rules that are in the rule set of the baseline domain.
\end{itemize}

We use \bcfq{} as baseline as all of the rule sets are supersets of it. We also compare each half-\starcfq{} dataset to the corresponding \starcfq{} dataset. As \lcfq{} and \xcfq{} are also supersets of \ocfq{}, we additionally include the relative sizes of these datasets compared to \ocfq{}.

\section{Effect of Compound Divergence}
\label{app:divergence-comparison-with-prev-paper}

Table~\ref{tab:experiment-3.2} shows mean error rates observed for each train set size for compound divergence splits at various levels of compound divergence from our experiments performed on  \ucfq{}. This is the same information as in Figure~\ref{fig:experiment-3.2}, but in tabular form.

\begin{table*}[tb]
\begin{tabular}{@{}lllllllll@{}}
\hline
Comp.& & & & & & & & \\
div.& 10,314 & 20,627 & 30,940 & 61,879 & 103,132 & 309,395 & 618,790 & 1,031,270\\
\hline
\hline
0.0 & 0.014$_{\pm 0.001}$ & 0.008$_{\pm 0.000}$ & 0.007$_{\pm 0.000}$ & 0.004$_{\pm 0.000}$ & 0.004$_{\pm 0.000}$ & 0.005$_{\pm 0.000}$ & 0.005$_{\pm 0.000}$ & 0.005$_{\pm 0.000}$\\
0.1 & 0.232$_{\pm 0.004}$ & 0.188$_{\pm 0.004}$ & 0.166$_{\pm 0.006}$ & 0.147$_{\pm 0.007}$ & 0.130$_{\pm 0.007}$ & 0.036$_{\pm 0.005}$ & 0.021$_{\pm 0.003}$ & 0.018$_{\pm 0.002}$\\
0.2 & 0.583$_{\pm 0.009}$ & 0.542$_{\pm 0.009}$ & 0.519$_{\pm 0.008}$ & 0.394$_{\pm 0.012}$ & 0.322$_{\pm 0.013}$ & 0.133$_{\pm 0.013}$ & 0.066$_{\pm 0.009}$ & 0.040$_{\pm 0.009}$\\
0.3 & 0.755$_{\pm 0.009}$ & 0.729$_{\pm 0.010}$ & 0.724$_{\pm 0.008}$ & 0.605$_{\pm 0.017}$ & 0.533$_{\pm 0.025}$ & 0.225$_{\pm 0.019}$ & 0.173$_{\pm 0.021}$ & 0.110$_{\pm 0.015}$\\
0.4 & 0.838$_{\pm 0.007}$ & 0.807$_{\pm 0.008}$ & 0.787$_{\pm 0.011}$ & 0.715$_{\pm 0.017}$ & 0.607$_{\pm 0.021}$ & 0.358$_{\pm 0.026}$ & 0.287$_{\pm 0.029}$ & 0.224$_{\pm 0.021}$\\
0.5 & 0.945$_{\pm 0.007}$ & 0.915$_{\pm 0.008}$ & 0.886$_{\pm 0.008}$ & 0.808$_{\pm 0.010}$ & 0.769$_{\pm 0.012}$ & 0.531$_{\pm 0.023}$ & 0.453$_{\pm 0.030}$ & 0.329$_{\pm 0.035}$\\
0.6 & 0.979$_{\pm 0.003}$ & 0.950$_{\pm 0.007}$ & 0.906$_{\pm 0.010}$ & 0.885$_{\pm 0.011}$ & 0.861$_{\pm 0.016}$ & 0.699$_{\pm 0.021}$ & 0.582$_{\pm 0.031}$ & 0.436$_{\pm 0.033}$\\
\hline
\end{tabular}

\centering
    \caption{Effect of compound divergence on scalability curve. Error rate vs train set size for compound divergence splits. Same data that is shown in Figure~\ref{fig:experiment-3.2}. Shows mean error rate with 95\% confidence interval over 5 replicas of 36 splits for each compound divergence level and train set size.
    }
    \label{tab:experiment-3.2}
\end{table*}

\begin{figure}[tb]
   \centering \footnotesize
  \includegraphics[width=0.45\textwidth]{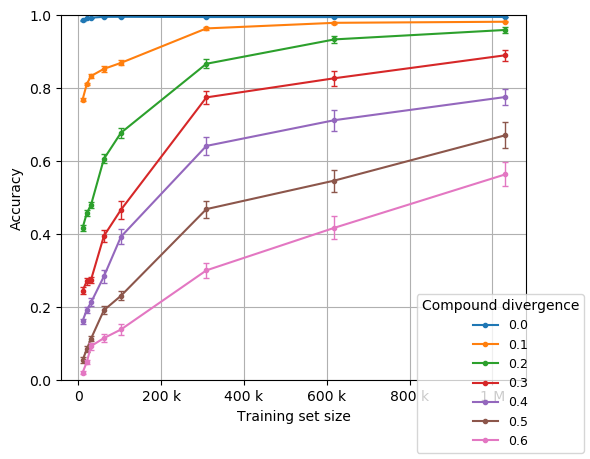}
   \caption{Effect of compound divergence on accuracy scalability curve. This is the same information as in Figure~\ref{fig:experiment-3.2} but reformatted for easier comparison with results in Figure 10 of \citet{keysers2020}.
   The differences between our experiments and those of \citet{keysers2020} are that our compound divergence splits are generated using as example pool a random sample of \ucfq{} that is 10x the size of \cfq{}, the Transformer is trained with 500k rather than 35k steps, and we average over a larger number of data points.
   }
   \label{fig:experiment-3.2-comparison-with-prev-paper}
\end{figure}

Figure~\ref{fig:experiment-3.2-comparison-with-prev-paper} 
 also 
shows the same information as in Figure~\ref{fig:experiment-3.2}, but plotting accuracy instead of the error rate, for easier comparison with the results obtained by \citet{keysers2020} in smaller scale experiments using \cfq{}.
Overall, our experiments show a similar trend to that seen in \citet{keysers2020}, but in a clearer form due to averaging over larger number of splits at each point and covering a wider range of training sizes. Note that the lower starting accuracies in our experiments are due to the tendency for accuracies at any given training size and compound divergence to be lower when the overall example pool is larger (an artifact of the compound divergence splitting algorithm).

\section{Blending of Augmented Train Set}
\label{app:blending-algorithm}

Here we describe the details of how we blend the in-domain train set $W_{TRAIN}$ with the supplementary train set $V_{TRAIN}$ in the experiments of Section~\ref{sect:effect-related-domain}.

As described in Section~\ref{sect:effect-related-domain}, we use two different setups for these experiments. In the ``equal-weighting'' setup of Section~\ref{sect:equal-weighting}, the in-domain portion of a train set is a fixed set of $n_{in}$ examples, and the supplementary portion varies in size ($n_{sup}$).
As no adjustment in sample weighting is done, the ratio of samples from the in-domain vs. supplementary sets varies in an identical manner.
In the ``over-weighting'' setup of Section~\ref{sect:over-weighting}, we generate train sets where
the number of unique examples from in-domain vs. supplementary sets varies in the same way as in the ``equal-weighting'' setup, but where examples are replicated as needed to keep
the ratio between in- and out-domain examples fixed at $1.0$.
We denote the number of unique examples selected from the in-domain pool in both cases by $n_{in}$. In the ``over-weighting'' experiments, this usually differs from the total number of examples that come from the in-domain pool, due to the replication.


In particular, we do the following:

\begin{enumerate}
    \item We take a pseudo-random set of examples from the question pool in the target domain (``in-domain pool'') that keeps the original distribution, and we split it to a train and a test set. To ensure that the test set is fixed across experiments of the same domain, we do the splitting by sorting the examples in the pool by their hashes and selecting the first $n_{test}$ examples as the test set and the next $n_{in}$ examples as the train set. In the ``equal-weighting'' experiments, if $n_{in} < n_{sup}$, we replicate the train set as many times as needed to get exactly $n_{sup}$ questions, truncating the last copy if $n_{sup}$ is not divisible by $n_{in}$. We call the train set obtained this way the ``in-domain train set''. As a result of this algorithm, a smaller in-domain train set is always a subset of a larger one in both setups.
    \item To make the results easier to interpret, we filter out from the pool of related examples (``supplementary pool'') all the examples that could be generated using the grammar of the target domain (and thus belong to that domain). This also removes the possibility of accidentally re-using a test question in the train set.
    \item We sort the filtered examples by their hashes and take the first $n_{sup}$ examples as the ``supplementary train set''.
    \item We take the union of the in-domain and supplementary train sets and use it as the train set in experiments.
\end{enumerate}

\section{Equal Weighting of Examples from Target and Related Domains: Detailed Results}
\label{app:equal-weighting-details}

\begin{figure*}[t!]
   \centering
   \includegraphics[width=\textwidth]{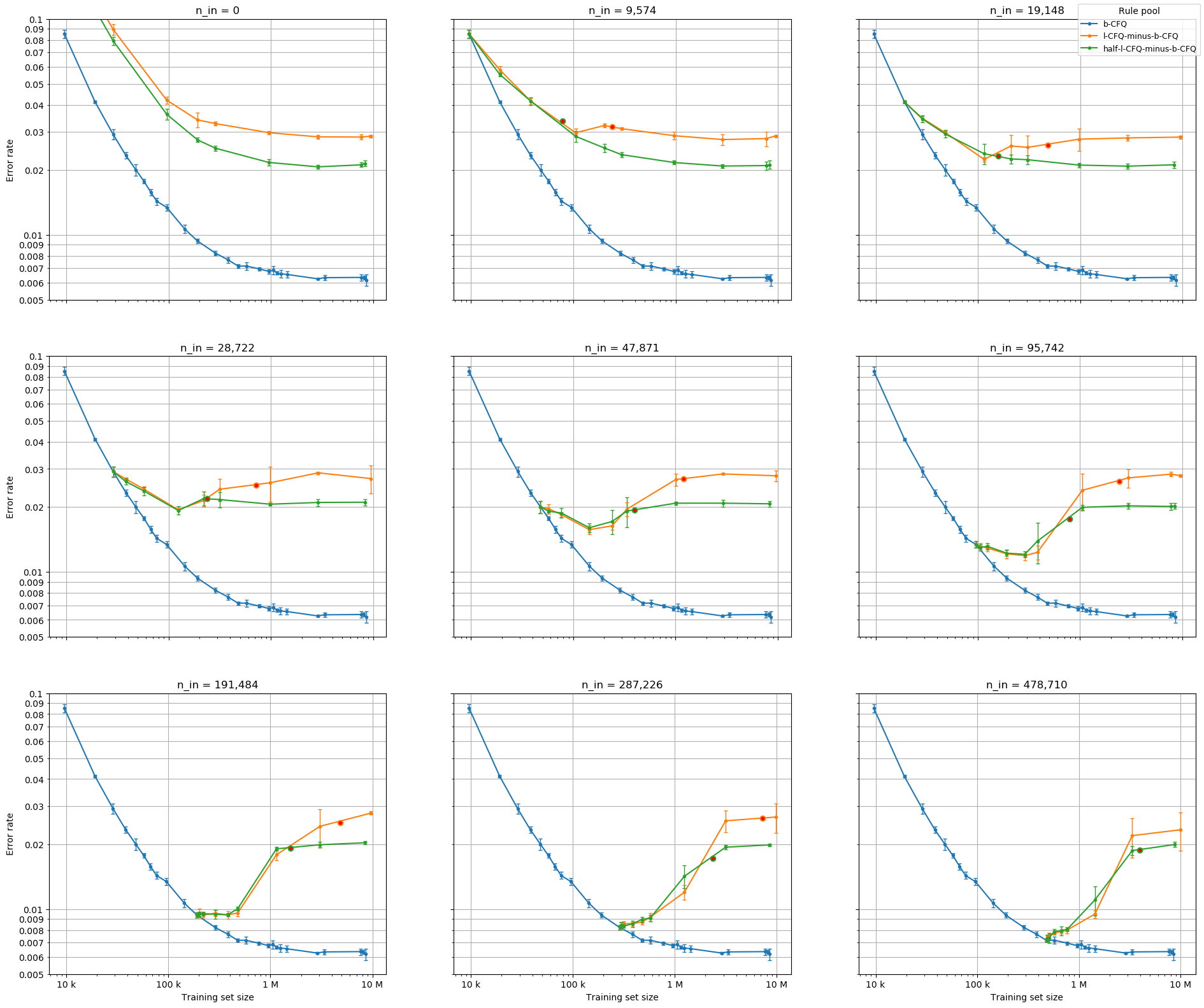}
   \caption{Error rate vs train set size, log-log scale, \textbf{leaf} changes (\bcfq{} test set, \bcfq{} + \lcfq{} train set), all initial sizes.}
   \label{fig:experiment-6.4-full}
\end{figure*}

\begin{figure*}[t!]
   \centering
   \includegraphics[width=\textwidth]{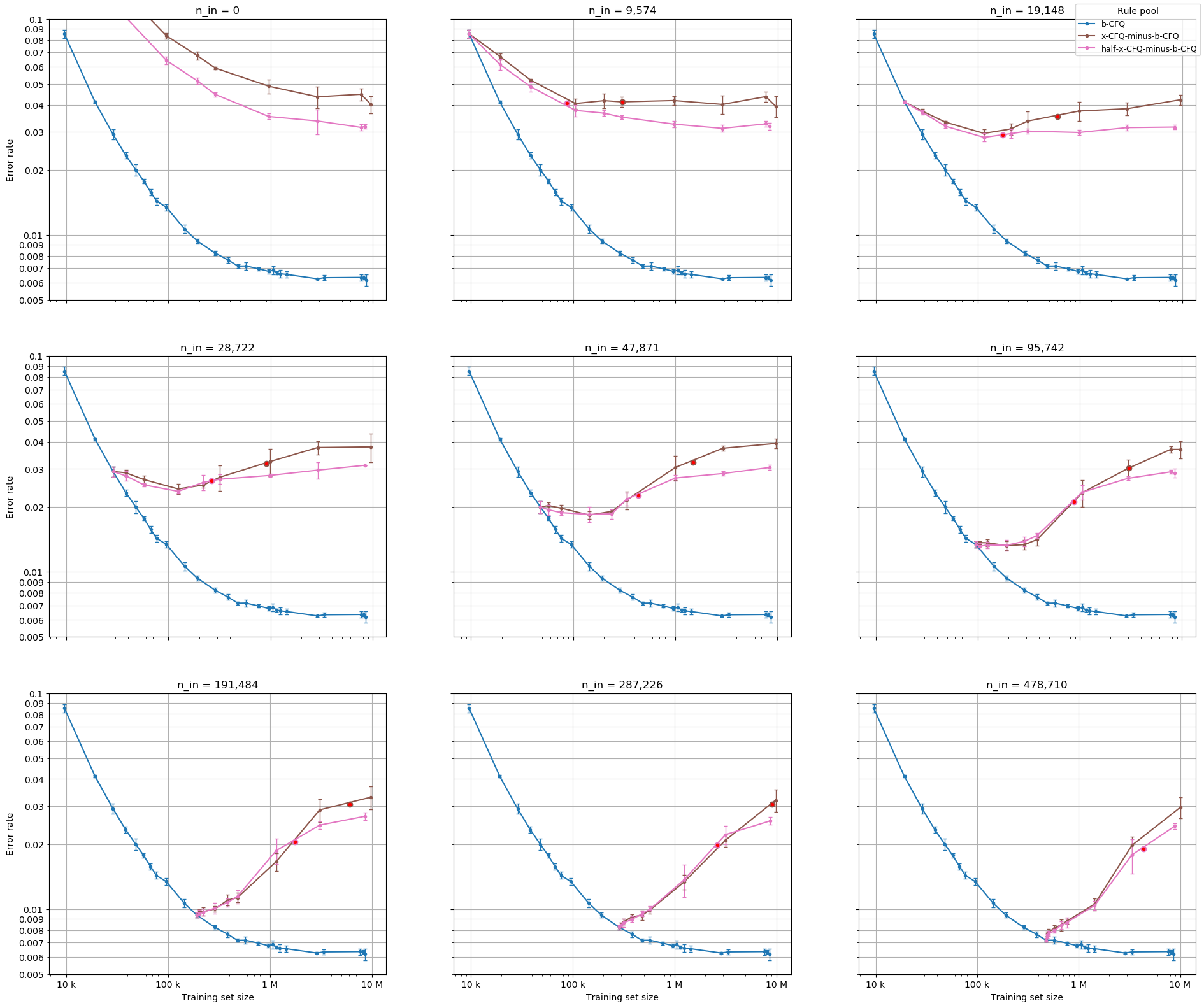}
   \caption{Error rate vs train set size, log-log scale, \textbf{leaf} and  \textbf{non-leaf} changes (\bcfq{} test set, \bcfq{} + \xcfq{} train set), all initial sizes.}
   \label{fig:experiment-8.1-full}
\end{figure*}

The complete results for experiments summarized in Section~\ref{sect:equal-weighting} are presented in Figures~\ref{fig:experiment-6.4-full} and \ref{fig:experiment-8.1-full}. Here the experiments are done with size of target domain data $n_{in}$ being $10k$, $20k$, $30k$, $50k$, $100k$, $200k$, $300k$, and $500k$. A curve with $n_{in} = 0$ (i.e., no target domain data) is also shown for comparison.

From Figure~\ref{fig:experiment-6.4-full} we can observe that in all cases a nearly identical plateau is reached as the distribution of the training set gets closer and closer to that of the supplementary domain $V$, regardless of the number of in-domain examples provided. Specifically, the half-\lcfq{} error rate stabilizes at roughly $0.03$, while for \lcfq{} the plateau is at value $0.02$. Note that these accuracies could be achieved equivalently by training with just $30k$ (resp. $50k$) training examples from \bcfq{} alone, without the benefit of any supplementary data. Also note that starting from $n_{in} = 200k$ adding related data does not improve error rate.

Figure~\ref{fig:experiment-8.1-full}, which represents data for half-\xcfq{} and \xcfq{}, demonstrates similar behaviour. On the graphs with smaller values of $n_{in}$ the error rate stabilizes at roughly $0.04$ (for half-\xcfq{}) and $0.03$ (for \xcfq{}). 



\section{Augmenting with Examples from \ncfq{} or half-\ncfq{}}
\label{app:blending-with-ncfq}

\begin{figure*}[t!]
   \centering
   \includegraphics[width=\textwidth]{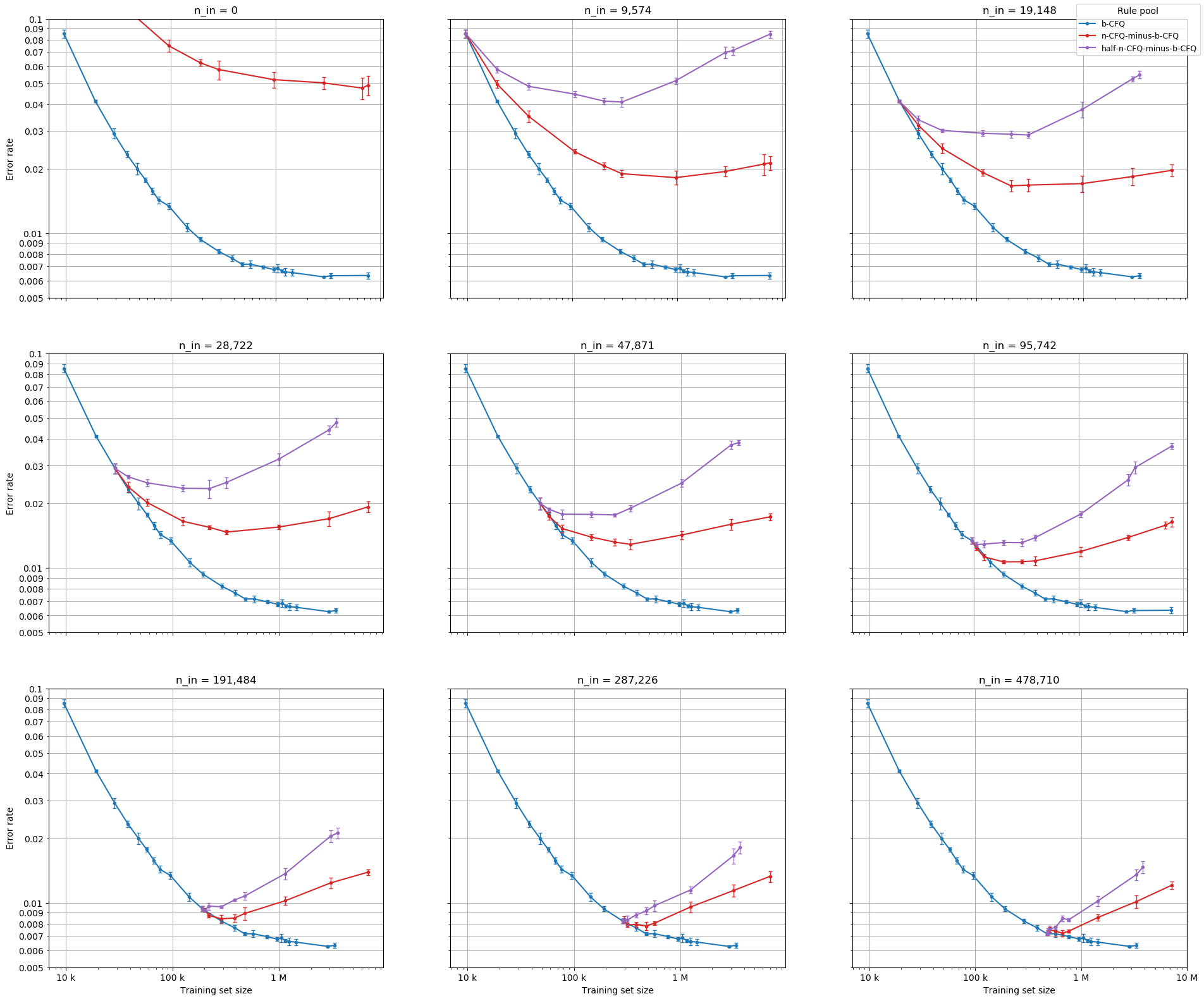}
   \caption{Error rate vs train set size, log-log scale, \textbf{non-leaf} changes (\bcfq{} test set, \bcfq{} + \ncfq{} train set), all initial sizes.}
   \label{fig:experiment-7.1-full}
\end{figure*}

Figure~\ref{fig:experiment-7.1-full} shows for completeness the results of experiments in mixing examples from a related domain, where the related domain is one that differs from the target domain only by the addition of non-leaf rules -- i.e., either \ncfq{} or half-\ncfq{}. The results from these experiments are unusual in that error rates when augmenting with half-\ncfq{} are even higher than when augmenting with \ncfq{}, contrary to the trend observed elsewhere that error rates drop more slowly the more distant the related domain is from the target domain.

We do not consider \ncfq{} or half-\ncfq{} to be well-suited to this particular experiment setup, however, due to the large overlap between these domains and the target domain \bcfq{}, and have accordingly omitted these results from the discussion in the main body. Specifically, as the \ncfq{} domain is only 1.28x the size of that of \bcfq{} (see Appendix~\ref{app:domain-size}), this means that the space of related domain examples excluding those that are part of the target domain is only 0.28x the size of the target domain. This effect is even more pronounced for half-\ncfq{}, whose domain is only 1.07x the size of that of \bcfq{}. This makes the results difficult to interpret, as the large fraction of \ncfq{} or half-\ncfq{} examples that are filtered out due to overlap with the \bcfq{} domain would magnify any potential side effects on the question distribution that might result from this filtering process, potentially to the point that this factor may dominate any observed changes in the error rate. We suspect this to be the primary reason for the observed higher error rates when augmenting with half-\ncfq{}, but would require deeper investigation to confirm.

\begin{figure*}[tb]
  \centering \footnotesize
  (a)\includegraphics[width=0.47\textwidth]{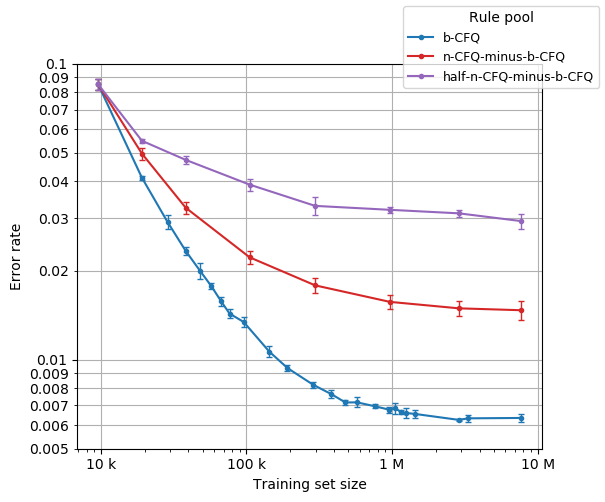}
  (b)\includegraphics[width=0.47\textwidth]{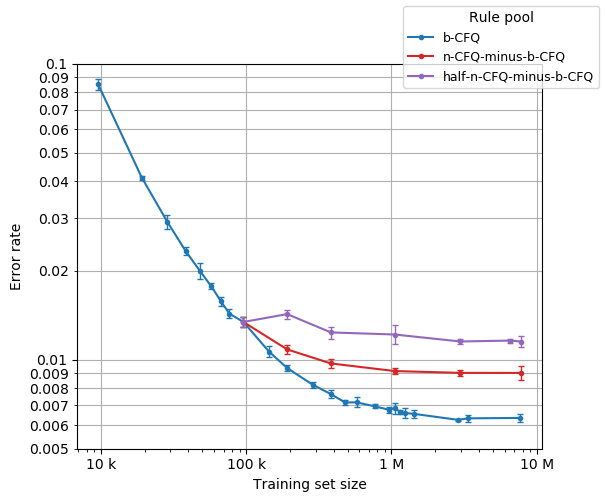}
  \caption{Error rate vs train set size, log-log scale, fixed ratio: \textbf{10k} \bcfq{} examples (a) and \textbf{100k} \bcfq{} examples (b), \bcfq{} test set, \bcfq{} + (\ncfq{} or half-\ncfq{}) train set.
  }
  \label{fig:experiment-7-ratio}
  \label{fig:experiment-7-ratio-small} 
  \label{fig:experiment-7-ratio-large} 
\end{figure*}

Figure~\ref{fig:experiment-7-ratio} is similar to Figure~\ref{fig:experiment-6-8-ratio} but demonstrates the effect of blending with \ncfq{} and half-\ncfq{}. 

\setcounter{topnumber}{5}
\setcounter{bottomnumber}{5}
\setcounter{totalnumber}{10}
\renewcommand{\topfraction}{0.9}
\renewcommand{\bottomfraction}{0.9}
\renewcommand{\textfraction}{0.1}
\renewcommand{\floatpagefraction}{0.9}


\end{document}